\documentclass{article} 
\usepackage{colm2024_conference}
\colmfinalcopy
\usepackage[inline]{enumitem}
\usepackage{microtype}
\usepackage{hyperref}
\usepackage{url}
\usepackage{booktabs}
\usepackage{graphicx}
\usepackage{placeins}
\usepackage{caption}
\usepackage{dingbat} 
\usepackage{subcaption}
\usepackage{amsmath}
\usepackage{booktabs}
\usepackage{siunitx}
\usepackage{wrapfig}
\newcolumntype{d}{S[
    input-open-uncertainty=,
    input-close-uncertainty=,
    parse-numbers = false,
    table-align-text-pre=false,
    table-align-text-post=false
 ]}

\newcolumntype{P}[1]{>{\centering\arraybackslash}p{#1}}

\title{
Secret Keepers: The Impact of LLMs on Linguistic Markers of Personal Traits
}

\author{Zhivar Sourati$^\diamond$, 
  Meltem Ozcan$^\dagger$,
  Colin McDaniel$^\dagger$,
  Alireza Ziabari$^\diamond$,
  \ \\
  \textbf{Nuan Wen}$^\diamond$, 
  \textbf{Ala Tak}$^\diamond$,
  \textbf{Fred Morstatter}$^\diamond$,
  \&
  \textbf{Morteza Dehghani}$^{\diamond\dagger}$ \\
$^\diamond$Department of Computer Science, University of Southern California,\\ 
$^\dagger$Department of Psychology, University of Southern California,\\ Los Angeles, CA, USA \\
\texttt{\{souratih,ozcan,cbmcdani,salkhord,}\\\texttt{nuanwen,nekouvag,morstatt,mdehghan\}@usc.edu} \\
}

\begin{document}

\maketitle

\begin{abstract}
Prior research has established associations between individuals' language usage and their personal traits; our linguistic patterns reveal information about our personalities, emotional states, and beliefs. However, with the increasing adoption of Large Language Models (LLMs) as writing assistants in everyday writing, a critical question emerges: are authors' linguistic patterns still predictive of their personal traits when LLMs are involved in the writing process? We investigate the impact of LLMs on the linguistic markers of demographic and psychological traits, specifically examining three LLMs — GPT3.5, Llama 2, and Gemini — across six different traits: gender, age, political affiliation, personality, empathy, and morality. Our findings indicate that although the use of LLMs slightly reduces the predictive power of linguistic patterns over authors' personal traits, the significant changes are infrequent, and the use of LLMs does not fully diminish the predictive power of authors' linguistic patterns over their personal traits. We also note that some theoretically established lexical-based linguistic markers lose their reliability as predictors when LLMs are used in the writing process. Our findings have important implications for the study of linguistic markers of personal traits in the age of LLMs.
\end{abstract}

\section{Introduction}
\label{sec:intro}

Linguistic patterns, such as semantic, lexical, and stylistic features of natural language, carry meaningful information about their authors \citep{pennebaker2011secret}. For instance, \citet{pennebaker1999linguistic} gathered essays written by 1,203 psychology students and captured their personality traits separately using questionnaires. By studying the link between linguistic patterns in the essays and the authors' personalities, they found associations such as a positive correlation between neuroticism and the use of negative emotion words. Similar linguistic markers that are predictive of authors' personal traits have been found for other psychological constructs and demographic attributes \citep[e.g.,][]{oberlander2006language, moreno2021can,mairesse2007using, schwartz2013personality}. It is noteworthy that these markers are often difficult to consciously conceal, making them valuable tools for psychological analysis and inferring otherwise private information from linguistic patterns. 



More recently, Large Language Models (LLMs), especially those with easy-to-use public interfaces, have emerged to assist in a wide range of writing tasks, including composing emails, social media posts, or essays, leading individuals to increasingly rely on them as writing assistants in their daily lives \citep{sakirin2023user,alafnan2023chatgpt,haynes2023new}. Such interactions between LLMs and their users have spurred researchers to emphasize the necessity for gaining a deeper understanding of the inner characteristics of LLMs \citep{safdari2023personality,caron2023manipulating,abdulhai2023moral} and the social consequences of their rapid adoption \citep{wu2023brief}. However, to our knowledge, no existing research has investigated the impact of LLMs on the linguistic patterns of authors, and the information embedded in these patterns about authors' personal traits, when LLMs are involved in the writing process. 



In this study, we aim to address this gap in the literature and, more concretely, answer the question \textit{do LLMs preserve linguistic markers of authors' attributes?} We rely on data sources that contain authors' written texts alongside certain demographic and psychological attributes about authors through self-reports (see \autoref{sec:psychological-constructs}). Specifically, we focus on three psychological constructs on which individuals differ, i.e., personality, dispositional empathy, and moral values, and three demographic attributes, i.e., gender, age group, and political affiliation.
We then ask LLMs to generate variations of the original texts, covering three LLMs: GPT3.5 \citep{chatgpt}, Llama 2 \citep{llama}, and Gemini \citep{gemini}, allowing us to compare their impacts (see \autoref{subsec:rewriting-essays}). 
In \autoref{sec:evaluation-framework}, first, we assess the predictive power of linguistic patterns regarding authors' attributes inherent in the original texts and compare them with the texts generated by LLMs. We then ground our analysis in established fine-grained linguistic markers of personal traits and explore the impact of LLMs on these theoretically grounded markers.

Our results suggest that, overall, there is a slight decrease in the predictive power of authors' linguistic patterns over their personal traits when LLMs are involved in the writing process. However, significant declines across all possible choices of classifiers, LLMs, and prompts are infrequent. Further, our investigations demonstrate that the fine-grained lexical-based markers of authors' attributes are altered in some cases between authors' original texts and LLM-generated texts. The observed changes in these fine-grained linguistic markers partially explain the aforementioned decline in predictive power, prompting questions about the reliability of specific linguistic markers of personal traits in the age of LLMs. 

\section{Related Work}
\label{sec:related-work}

\paragraph{Linguistic markers of demographic and psychological constructs.}
Numerous studies have demonstrated that authors' linguistic patterns provide information about their characteristics such as personality \citep[e.g.,][]{pennebaker1999linguistic,sun2018personality}, dispositional empathy \citep[e.g.,][]{buechel2018modeling,barriere2023findings} morality \citep[e.g.,][]{kennedy2021moral,ziabari2024reinforced}, gender \citep[e.g.,][]{cheng2011author,peersman2011predicting}, and political affiliation \citep[e.g.,][]{santurkar2023whose,baly2020we}. 
Inquiries into the relationship between language use and author attributes can be grouped into two approaches, each serving a different purpose \citep{kennedy2022handbook}: bottom-up approaches and top-down approaches. Bottom-up approaches support data-driven discovery of linguistic features for the purpose of predicting authors' attributes by finding any informative signal in people's speech or written text \citep[e.g.,][]{peersman2011predicting}.
Meanwhile, top-down approaches study the hypothesized fine-grained linguistic markers of authors' attributes using well-curated lexicons and statistical modeling with the goal of facilitating inference \citep[e.g.,][]{mairesse2007using}. 
This study aims to determine if established associations between language usage and authors' attributes remain meaningful when LLMs are used in writing.

\paragraph{LLMs \& social psychological research.}
Recent advancements in LLMs have opened new avenues in social psychological research. LLMs, renowned for their ability to generate human-like text \citep{herbold2023large}, offer unique tools for studying social interactions and provide valuable insights into human behavior and social dynamics \citep[e.g.,][]{park2023generative}. For instance, \citet{rao2023can} apply LLMs to detect personality traits from input text, while \citet{safdari2023personality} provide personality measurements for LLM-generated text. Other works explore the demographic-related and psychological features of LLMs themselves, such as personality \citep[e.g.,][]{miotto2022gpt}, world views and beliefs \citep[e.g.,][]{he2024whose}, creativity \citep[e.g.,][]{uludag2023testing}, and ability to empathize \citep[e.g.,][]{belkhir-sadat-2023-beyond}. While the human-like qualities of LLMs have garnered praise, numerous studies have highlighted significant limitations and biases in these models~\citep[e.g.,][]{abdurahman2023perils,messeri2024artificial}. Nevertheless, to our knowledge, the influence of LLMs on the linguistic markers of individuals' private attributes is yet to be explored. We aim to address this gap in the present study.

\section{Experimental Setup}
To study the influence of LLMs on the linguistic markers of authors' personal traits when LLMs are involved in writing, we first delve deeper into the particular personal traits we investigate, in \autoref{sec:psychological-constructs}. In \autoref{subsec:rewriting-essays}, we provide the details of how we use LLMs to generate variations of authors' original texts. Finally, in \autoref{sec:evaluation-framework}, we discuss how the authors' linguistic patterns that are predictive of their personal traits are studied and compared between authors' original and LLM-generated texts. 

\subsection{Author Attributes}
\label{sec:psychological-constructs}
For our investigation, we choose six well-researched demographic and psychological constructs: gender, age, political affiliation, personality, dispositional empathy, and morality. For each of these attributes, we use a dataset containing written text as well as the corresponding value of the attribute from the texts' respective author. Authors' demographic attributes were gathered using external sources (e.g. Wikipedia), while other psychological attributes were compiled using validated self-report questionnaires. Crucially, the texts and the individuals' demographic/psychological attributes were collected independently (i.e., author attributes were not annotated based on the written text).

\paragraph{Demographic attributes.} Among different demographic attributes, we consider age, gender, and political affiliation. As our data source, we use the United States Congressional Records \citep{gentzkow2018congressional}, which contains congressional floor speeches along with the speakers' demographic details \citep{fivethirtyeight_age_problem}. 
The dataset contains data from 8,520 speakers from the 43$^{\text{rd}}$ to the 114$^{\text{th}}$ Congress, having between 1 to 21,142 speeches with varying word lengths. As a filtering step, we begin with the speakers' longest utterances, selecting enough to reach a total of 4,000 words for each speaker. This is to remove short, uninformative utterances and to take into account the computational limits on the processed maximum number of tokens. We then sample the largest possible subset of speakers that would have a balanced number of males/females (due to the binary scheme of the data), Republicans/Democrats, and age groups of (27-40), (41-55), (56-70), and (over 70), totaling 710 speakers.

\paragraph{Personality.} 
To study personality, we choose the widely accepted Big Five personality model \citep{goldberg2013alternative,goldberg1990standard}, which describes five fundamental dimensions of personality: openness (OPN), conscientiousness (CON), extraversion (EXT), agreeableness (AGR), and neuroticism (NEU). We use the extended Essays dataset \citep{pennebaker1999linguistic} that contains 2,348 essays, each from a unique author, and the authors' scores on the 44-item Big-5 Personality Inventory \citep{john1991big}. Similar to \citet{celli2013workshop}, the labels are nominal classes with a median split (low/high) that are converted from $z$-scores of the actual numerical self-assessments. The authors were tasked with writing an essay using a stream-of-consciousness approach, encouraging them to think freely and record whatever thoughts came to mind. The authors then completed the personality assessment.

\paragraph{Dispositional empathy.} 
We use the Interpersonal Reactivity Index (IRI; \citealp{davis1980multidimensional}), which is one of the most common ways to assess individual differences in this construct. The IRI is a 28-item self-report questionnaire designed to measure empathy in adults, consisting of four dimensions: perspective-taking (PT), fantasy (FS), empathetic concern (EC), and personal distress (PD). For the purposes of our investigation, we use the Empathetic Conversations dataset \citep{omitaomu2022empathic}, previously used in the WASSA 2023 shared task \citep{barriere2023findings}, which contains essays written in response to news articles and IRI scores for the authors. Similar to the process employed for the Essays dataset, labels are nominal classes indicating the low/high level of each IRI dimension. After cleaning and preprocessing, the dataset contains data from 57 authors who wrote between 1 and 72 reaction essays, resulting in a total of 711 essays. To have an aggregated picture, we focus on concatenated essays from each author. Additionally, we explore how considering each essay as an independent observation, ignoring that certain texts are from the same author, impacts findings compared to the concatenated approach. 

\paragraph{Morality.} 
One of the most common theoretical frameworks of morality is Moral Foundation Theory (MFT; \citealp{graham2013moral,haidt2004intuitive,atari2023morality}), which proposes five innate and universally available psychological systems that people use to process moral matters: care/harm, fairness/cheating, loyalty/betrayal, authority/subversion, and purity/degradation. We use the YourMorals dataset \citep{kennedy2021moral} that contains 107,798 Facebook posts from 2,691 users alongside the authors' scores on the Moral Foundations Questionnaire \citep{graham2008moral}. Similar to the process employed for the Essays dataset, labels are nominal classes indicating the low/high level of each MFT dimension. Also, similar to the Empathetic Conversations dataset, we focus on the concatenated posts from each author. 

\subsection{LLMs' Variations of Authors' Texts}
\label{subsec:rewriting-essays}
To gather the LLMs' variations of authors' original texts, we employ a natural procedure that people might realistically engage in when utilizing LLMs, by prompting LLMs to rewrite a piece of text without pointing out any particular characteristic that the author might have or might want to integrate into the text. We chose the following two neutral prompts, one that limits changes to a bare minimum of syntactical revisions, and another that allows LLMs to freely make changes to the text:
\begin{itemize}
    \item Syntax\_Grammar (SG): 'Rewrite the following text using the best syntax and grammar and other revisions that are necessary: \{TEXT\}'
    \item Rephrase (R): 'Rephrase the following text: \{TEXT\}'
\end{itemize}

We utilize GPT3.5 \citep{chatgpt}, Llama 2 70B \citep{llama}, and Gemini pro \citep{gemini} with a deterministic generation setting (Temperature = 0) for our experiments. We focus on these three LLMs due to their popularity and easy-to-use public interfaces that enable the general public to use them without needing any expertise, as opposed to other LLMs, which might be more suitable for academic audiences. 

\subsection{Predicting Authors' Personal Attributes with Linguistic Patterns}
\label{sec:evaluation-framework}
In this section, we discuss two approaches for predicting author attributes using their linguistic patterns, and further discuss how we compare these linguistic patterns between the authors' original and LLM-generated texts. Similar to the categorization done by \citet{kennedy2022handbook}, we study authors' linguistic patterns predictive of their personal traits in two ways:
\begin{enumerate*}
    \item \textbf{Bottom-up} (data-driven) approach that relies on data-driven methodology to find linguistic cues that are predictive of the authors' attributes;
    \item \textbf{Top-down} (theory-driven) approach that is grounded in psychological theory and focuses on fine-grained associations between predetermined lexical cues and authors' characteristics. 
\end{enumerate*}

\paragraph{Bottom-up (data-driven)  analysis.}
In this approach, we train classifiers that predict authors' attributes given their original texts. We then compare the predictive power of classifiers on the original and their corresponding LLM-generated held-out test splits. However, different classifiers with different architectures may capture different features and achieve varying predictive powers, making them incomparable to one another. Since the primary question is whether the predictive power is preserved when an LLM is involved in the writing process, regardless of the family of classifier being utilized, we adopt an aggregated point of view and focus on the ratio of unchanged performances across different classifiers and featurization techniques. As classifiers, we use four different families of classifiers, i.e., Support Vector Machines (SVMs), Logistic Regression, Random Forest, and Gradient Boosting, on top of two featurization techniques (TF-IDF, and OpenAI text-embedding-ada-002 embeddings). We also include Longformer \citep{beltagy2020longformer}, a Transformer model \citep{vaswani2017attention} suitable for processing the long documents in our datasets. This way, we cover both traditional, conventional, as well as state-of-the-art methods. We separate the data into train, validation, and test splits with 5-fold cross-validation for training the models and track $F_1$-macro scores. The comparison between models' performance on original and LLM-generated data is done across 40 runs with different random seeds to ensure the robustness of the results. Focusing only on the runs with above random performance on original texts, we filtered out models for which there were fewer than $20$ runs. In all the analyses, given that we perform multiple comparisons across classifiers, prompts, and LLMs, significance thresholds were adjusted using a Bonferroni correction \citep{bonferroni1936teoria}.

\paragraph{Top-down (theory-driven) analysis.}
In the top-down analysis, we adopt a more granular approach and examine the LLMs' impact on associations previously demonstrated in the literature between lexical cues and demographic/psychological attributes. This approach allows us to glean which specific linguistic cues were affected by LLMs and helps us explain changes in predictive power in the bottom-up approach. More concretely, we: 
\begin{enumerate*}
    \item choose psychologically validated dictionaries of words associated with authors' traits (or traits' dimension) of interest;
    \item for each dictionary category, calculate the ratio of the number of words belonging to that category in authors' texts (i.e., word frequency) to the total number of words in the texts as a standardized score;
    \item for continuous author traits (i.e., age, personality, empathy, morality), compute the Pearson correlation ($r$) between these standardized word frequencies and the $z$-scores of the traits of interest, and for categorical traits (i.e., gender and political affiliation), conduct a $t$-test to determine significant differences of scores across groups; and
    \item perform Bonferroni corrections to account for multiple comparisons across dimensions (for psychological constructs) and dictionary categories (i.e., dividing the significance threshold of 0.05 by the number of dimensions in the psychological construct and by the number of dictionary categories, e.g., for personality, which has five dimensions and 24 dictionary categories examined, the threshold is $0.05/(5 \cdot 24) = 0.000417$). 
\end{enumerate*}

As a general-purpose dictionary, we use the Linguistic Inquiry and Word Count (LIWC; version 22 and 07, merged; \citealp{pennebaker2022linguistic}), which covers various psychological and topical categories and social, cognitive, and affective processes. We select a collection of LIWC features for each author attribute based on theoretically backed hypotheses regarding which features would correlate best with the demographic/psychological trait of interest. In addition, we use content-specific dictionaries to conduct a more targeted investigation of the linguistic cues associated with specific psychological traits. Namely, we use the 10-category NRC Emotion Lexicon \citep{Mohammad13,mohammad-turney-2010-emotions} for associations with personality, the lexicon gathered by \citet{sedoc2019learning}, median split into high- and low-empathy and distress words (four categories) for associations with dispositional empathy, and the Moral Foundations Dictionary 2 (MFD2; \citealp{frimer2019moral}) for associations with moral foundations, which we test within each moral foundation (two categories per foundation).

\section{Results}
\label{sec:results}

\subsection{Do LLMs Change the Semantics in People's Writings?}
\label{subsec:overall-changes-in-texts}

\begin{figure}[h]
    \centering
    \includegraphics[width=\textwidth]{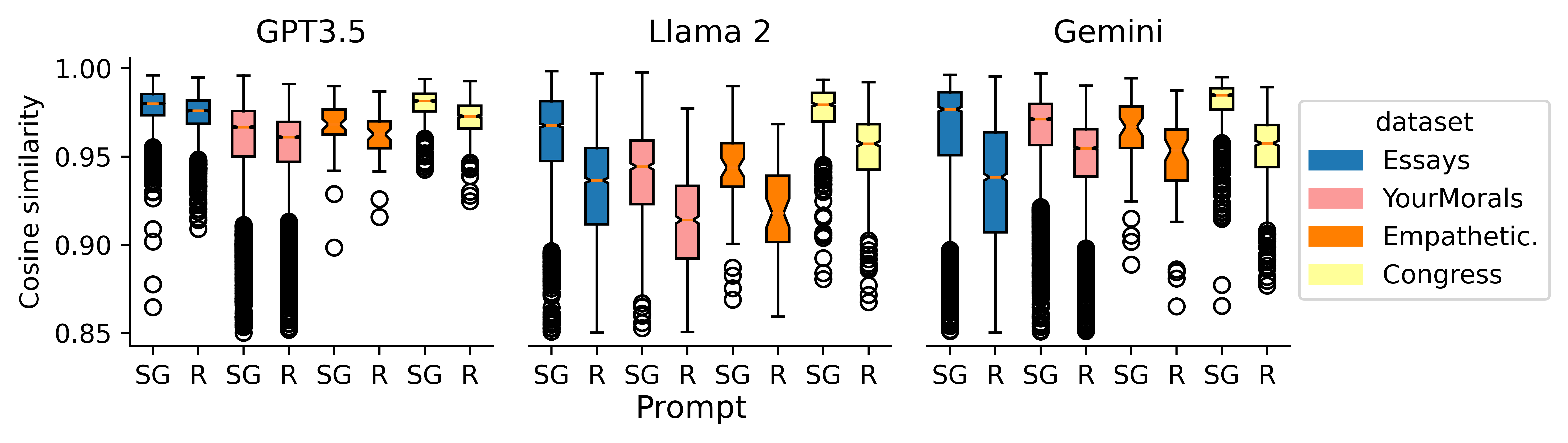}
    \vspace{-2em}
    \caption{Semantic similarity between original and LLM-generated texts (with \textbf{R}ephrase and \textbf{S}yntax\_\textbf{G}rammar prompts) across different data sources and utilized LLMs.}
    \label{fig:semantic-similarities-all}
\end{figure}

Texts generated by LLMs can differ from authors' original texts in terms of various syntactic and semantic linguistic patterns. We expected that LLMs would not change the meaning of original texts when neutral prompts are used in the rewriting process. \autoref{fig:semantic-similarities-all} demonstrates the semantic similarity between original and LLM-generated texts, computed as the cosine similarity between OpenAI text-embedding-ada-002 embeddings of the two text pieces. 

Aligned with our expectations, the semantic similarity between original and LLM-generated texts across all data sources, LLMs, and prompts was high ($75\%$ of scores were above $0.94$; see \autoref{tab:similarity_descriptive} in Appendix for descriptive statistics).
A Kruskal-Wallis (H) test, a non-parametric alternative to ANOVA,~\footnote{The Fligner-Killeen's test, which is robust against non-normality and large sample sizes, revealed that the assumption of homogeneity of variance was violated ($F = 5532.7$, $p < .001$).} was performed to compare similarity scores across LLMs and datasets, while a Mann–Whitney (U) test was used to compare the similarities across two prompts. Our results indicated that computed similarities were significantly different across LLMs ($H = 2718$, $p < .001, \eta^2=0.08$ [moderate effect size]), two prompt conditions ($U = 92224114$, $p<.001, r = 0.32$ [medium effect size]), as well as datasets ($H = 1758.5$, $p < .001, \eta^2=0.05$ [small effect size]).

Pairwise post-hoc Dunn's tests \citep{Dunn}\footnote{Dunn's tests are conventionally used for pairwise comparisons after a Kruskal-Wallis test is rejected.} with Benjamini-Hochberg corrections for multiple comparisons showed significant differences between all LLM pairs ($p < .001$; see \autoref{tab:pairwise_sim} in Appendix), with GPT3.5 being the most preservative LLM and Llama 2 being the least preservative LLM in terms of the level of semantic preservation when rewriting. With a similar post-hoc analysis, comparing the two prompt conditions, the Syntax\_Grammar prompt was more preservative than the Rephrase prompt with median similarity values of 0.97 and 0.96, respectively (see \autoref{tab:sample-original-rewritten-text-gpt} for an example of the original and LLM-generated texts). Finally, underscoring the role of context when studying LLMs' impact on authors' texts, our pairwise post-hoc tests showed significant differences between all dataset pairs aside from YourMorals vs. Empathetic Conversations (see \autoref{tab:pairwise_sim_datasets} in Appendix), which showed the highest level of semantic preservation for the Congress dataset and the least level of preservation for the YourMorals dataset.

\begin{table*}[h]
\centering
\scriptsize
\begin{tabular}{p{1.1cm}p{12cm}}
\toprule
Prompt & Text \\
\midrule
Original & it has been raining for the past 10 days and I guess that has a lot to do with everyone's attitudes. I feel a little relieved and tired though I had my first test of the semester this morning. I feel tired because I spent the majority part of the weekend partying rather then studying like I should have $\cdots$ \\
\midrule
Rephrase & It has been raining continuously, which seems to be affecting everyone's attitudes. Despite feeling relieved, I am also tired because I had my first test of the semester this morning. Unfortunately, I spent most of the weekend partying instead of studying, so $\cdots$ \\
\midrule
Syntax Grammar & It has been raining for the past few days, and I believe this weather has had a significant impact on everyone's attitudes. Personally, I feel a mixture of relief and fatigue, as I had my first test of the semester this morning. The tiredness stems from the fact that I spent the majority of the weekend partying instead of studying, as I $\cdots$ \\
\bottomrule
\end{tabular}
\caption{Variations of a sample text from the Essays dataset.}
\label{tab:sample-original-rewritten-text-gpt}
\end{table*}

\subsection{Do LLMs Wash Away Linguistic Markers of Author Attributes?}
\label{do-llms-wash-away-stuff}

\begin{figure}[h]
    \centering
    \includegraphics[width=\columnwidth]{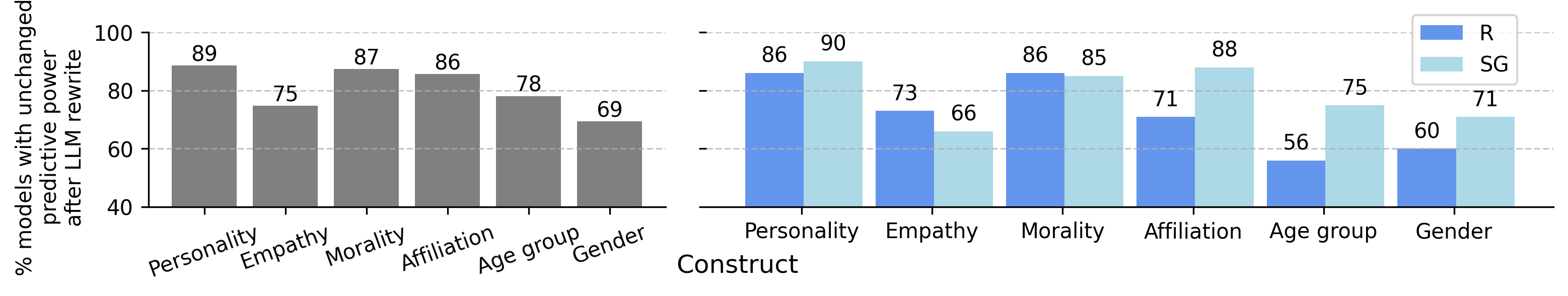}
    \caption{The ratio of classifiers with unchanged predictive power after LLM rewrites across different author attributes. The left plot shows the aggregated view and the right plot shows the variability across two prompts: \textbf{R}ephrase and \textbf{S}yntax\_\textbf{G}rammar.}
    \label{fig:overall-results-for-changes-in-predictive-power}
\end{figure}

Regardless of how LLMs change the style or meaning of authors' written text, the core question is whether LLMs preserve the linguistic patterns of authors that are predictive of the authors' characteristics. We observed that the performance of classifiers predicting authors' personal traits, on the LLM-generated texts was significantly lower compared to the performance on the original texts, across different constructs ($p<.001$; [small] Cohen's $d$ effect size, using a paired $t$-test; see \autoref{tab:ttest_res} in Appendix). Although we observed a 6\% decline in the absolute $F_1$ score of classifiers on average, the classifiers' performance did not go below random performance. Further, comparing the classifiers' performance on original and LLM-generated data, across all possible LLMs, prompts, or used classifiers, our results suggested that significant performance declines are infrequent. \autoref{fig:overall-results-for-changes-in-predictive-power}-left shows the ratio of classifiers with unchanged predictive power (using a paired $t$-test and a Bonferroni correction for multiple comparisons) after LLM rewrites, across different author attributes. With this more precise analysis that tries to address the main question of this study, we found that the predictive power of authors' linguistic patterns over their personal traits, although slightly reduced, is not fully diminished. LLMs' impact on the predictive power of linguistic markers of gender and dispositional empathy was higher, where the predictive power of $31\%$ and $25\%$ of classifiers decreased, respectively. The few occurrences where the predictive power of classifiers significantly changed were mostly using logistic regression and TF-IDF as classifiers. 

Upon closer investigation, predictive power was more often reduced rather than increased, with the exception of the Care dimension of morality, where the involvement of LLMs amplified the predictive power of linguistic patterns in predicting the level of the authors' Care, which we conjecture to be a byproduct of the reinforcement learning from human feedback (RLHF) stage of LLMs' fine-tuning, promoting certain behaviors. See \autoref{fig:results-for-differences-for-all-labels} in Appendix for a more fine-grained illustration of the changes in the predictive power per dimension of authors' personal traits.

\begin{figure}[h]
    \centering
    \includegraphics[width=\columnwidth]{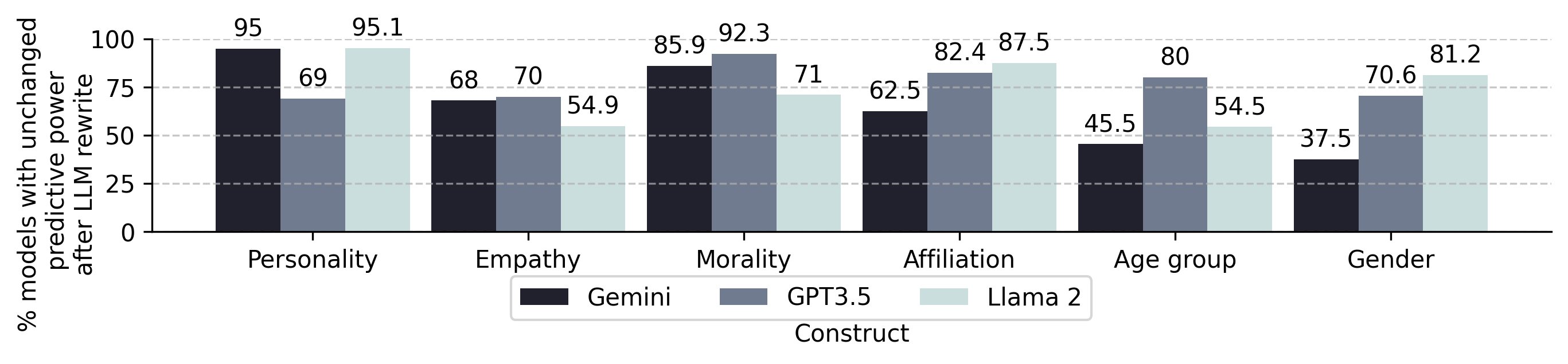}
    \caption{The ratio of models with unchanged predictive power after LLM rewrite, across different author attributes and different LLMs.}
    \label{fig:difference-between-llms}
\end{figure}

LLMs having different training pipelines, the impact of different LLMs on the predictive power of linguistic patterns over author attributes might differ. \autoref{fig:difference-between-llms} demonstrates the ratio of classifiers with unchanged predictive power across different LLMs. Although jointly across all author attributes, we did not find a significant difference in impact between LLMs, when we only focused on the demographic attributes (i.e., gender, age, and political affiliation), the difference between LLMs' impact was significant (Cochran's Q test with $\chi^2 = 12.6$, $p = .001$). We found that the level of preserved predictive power by Gemini was lower than GPT3.5 ($\text{McNemar's}\;\chi^2 = 9.0, p = 0.01, \text{Odds ratio} = 0.14$ [large effect size]) and Llama 2 ($\chi^2 = 8.89, p = 0.01; \text{Odds ratio} = 0.18$ [large effect size]), while the impact of Llama 2 and GPT3.5 was not significantly different.

Moreover, our results suggested that SG prompt is more preservative than the R prompt ($\text{McNemar's}\;\chi^2 = 3.76, p = 0.05, \text{Odds ratio} = 0.58$ [small effect size]) in keeping the same predictive power of linguistic patterns over author attributes, which underscores the role of prompts in the LLMs' impact on linguistic markers of personal traits (see \autoref{fig:overall-results-for-changes-in-predictive-power}-right), with stronger effects in the case of gender, age group, and political affiliation prediction. 


\begin{figure*}[ht]
    \centering
    \includegraphics[width=\textwidth]{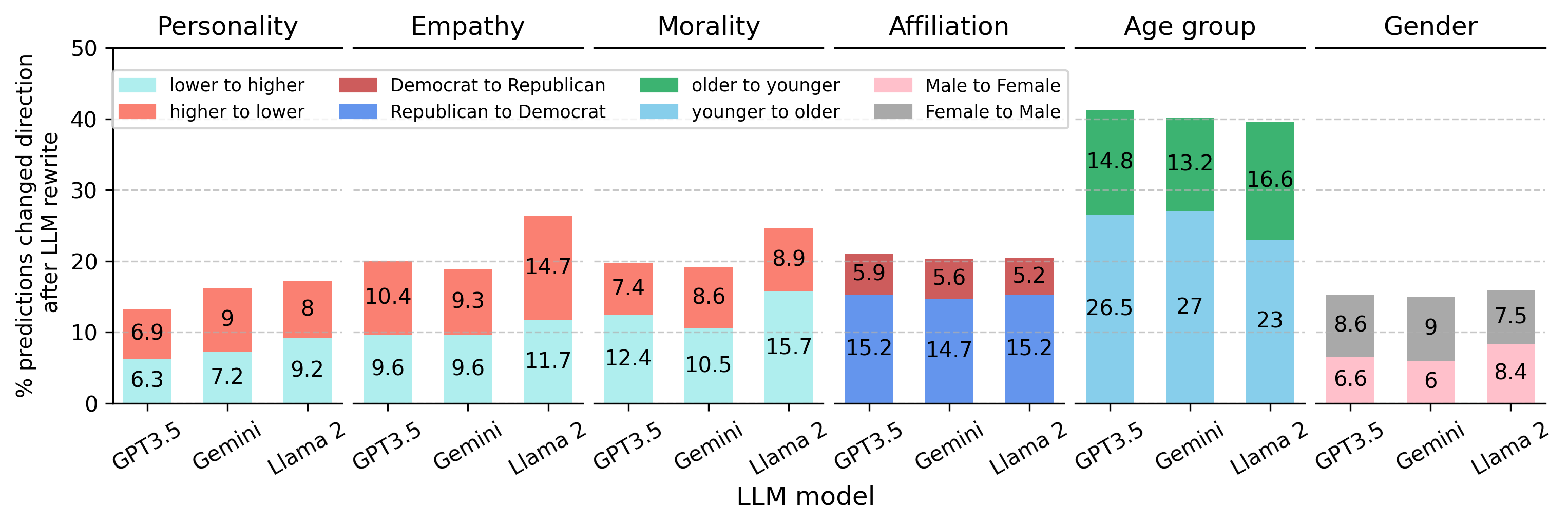}
    \caption{Ratios of correct author attribute predictions on original texts that had different predicted labels on LLM-generated texts, grouped by the direction of change in predictions.}
    \label{fig:difference-in-predictions-agg}
\end{figure*}

Motivated by research identifying specific human-like traits of LLMs (e.g., whether an LLM is introverted or extroverted; \citealp{santurkar2023whose,miotto2022gpt}), we analyzed the direction in which LLMs change the predictions of the classifiers, which can be a proxy for what LLMs promote in the generated text. \autoref{fig:difference-in-predictions-agg} demonstrates the ratio of predictions that changed from correct to incorrect, grouped by the direction of this change across different author attributes and LLMs. Reflected by more changes in the 'higher to lower' direction, our results suggested that LLM-generated text had levels of empathy that are associated with authors having lower levels of empathy compared to the actual authors ($t(6698) = 20.18, p <.001, \text{Cohen's}\;d = 0.35$ [small effect size]; using a paired $t$-test). With a similar reasoning, LLM-generated text was associated with authors having higher levels of morality ($t(3051) = 21.25, p <.001, d = 0.61$ [medium effect size]) and higher age groups ($t(597) = 16.84, p <.001, d = 1.09$ [large effect size]) compared to the actual authors. LLM-generated text confused classifiers to incorrectly predict authors as being Democrats more often than confusing them the other way around ($t(867) = 24.89, p <.001, d = 1.35$ [large effect size]). Finally, Gemini and GPT3.5 involvement was accompanied by more 'Female to Male' prediction changes ($t(613) = 5.68, p <.001, d = 0.37$ [small effect size]), while Llama 2 involvement changed gender predictions partially equally in both directions. For a more detailed discussion, refer to \autoref{appendix:What-Author-Attributes-do-LLMs-Promote-the-Most?}.


\subsection{Do LLMs Change Fine-Grained Associations Between Lexical Cues and Authors’ Attributes?}
\label{do-llms-change-the-associations}

\begin{table}[]
\begin{minipage}[b]{.50\textwidth}

\centering
\resizebox{0.98\textwidth}{!}{%
\begin{tabular}{llc|ccc|ccc}
&& \multicolumn{1}{c}{ }& \multicolumn{3}{c}{Rephrase} & \multicolumn{3}{c}{Syntax-Grammar} \\
 \cmidrule(lr){4-6}\cmidrule(l){7-9}

\multicolumn{2}{c}{\textbf{Personality}} & \multicolumn{1}{c}{Original}& \multicolumn{1}{c}{Gemini}& \multicolumn{1}{c}{GPT3.5} &\multicolumn{1}{c}{Llama 2} & \multicolumn{1}{c}{Gemini}& 
   \multicolumn{1}{c}{GPT3.5} & \multicolumn{1}{c}{Llama 2} \\
  \hline
OPN  &  drives &  & \checkmark & \checkmark& &  &  \checkmark &  \\ 
&  swear & \checkmark & & \checkmark& &  \checkmark&   & \checkmark \\ 
&  BigWords & \checkmark & & & &  & &   \\ 
   \hline
CON &anger emo & \checkmark & \checkmark &\checkmark  & &  & \checkmark & \checkmark   \\ 
&  swear & \checkmark & & \checkmark & & &  \checkmark &\checkmark    \\ 
  \hline
EXT & Social & \checkmark & & \checkmark & &  & &  \\ 
&  pos emo & \checkmark & \checkmark &  \checkmark & \checkmark & & \checkmark & \checkmark \\
  \hline
AGR &  anger emo & \checkmark & \checkmark& \checkmark& \checkmark& \checkmark & \checkmark& \checkmark\\ 
&  swear & \checkmark & & & &  & \checkmark&\checkmark   \\
   \hline
NEU & I pronoun & \checkmark & &  \checkmark&  \checkmark& \checkmark & \checkmark & \checkmark\\ 
&  affect &  &  \checkmark &  \checkmark & &   &   \checkmark & \\ 
&  neg emo & \checkmark & \checkmark& \checkmark & \checkmark& \checkmark &  \checkmark & \checkmark\\ 
& anger emo & \checkmark & & & & & & \\
\multicolumn{9}{c}{\vspace{0.2cm}}\\
\multicolumn{2}{c}{\textbf{Demographics}} & & & & & & & \\
\hline
gender & article & \checkmark & \checkmark & \checkmark & \checkmark & \checkmark & \checkmark & \checkmark \\
& anx emo & \checkmark & \checkmark & & & & & \\
& Social & \checkmark & \checkmark & \checkmark & & \checkmark & \checkmark & \checkmark \\
age & we & & \checkmark & & \checkmark& & \checkmark & \\
affiliation & adverb & \checkmark & & & \checkmark & \checkmark & \checkmark & \checkmark \\
\bottomrule
\end{tabular}}

\end{minipage}
\begin{minipage}[b]{.50\textwidth}

\centering
\resizebox{0.97\textwidth}{!}{%
\begin{tabular}{llc|ccc|ccc}
&& \multicolumn{1}{c}{ }& \multicolumn{3}{c}{Rephrase} & \multicolumn{3}{c}{Syntax-Grammar} \\
 \cmidrule(lr){4-6}\cmidrule(l){7-9}
\multicolumn{2}{c}{\textbf{Morality}} & \multicolumn{1}{c}{Original}& \multicolumn{1}{c}{Gemini}& \multicolumn{1}{c}{GPT3.5} &\multicolumn{1}{c}{Llama 2} & \multicolumn{1}{c}{Gemini}& 
\multicolumn{1}{c}{GPT3.5} & \multicolumn{1}{c}{Llama 2} \\
\hline
Fairness  &  pronoun &  &  & & & \checkmark  &   &  \\ 
&  religion & \checkmark & & & &  &   &  \\ 
   \hline
Loyalty & affiliation & \checkmark & \checkmark &\checkmark  & \checkmark&  \checkmark&  &   \\ 
&  family & \checkmark  & \checkmark & \checkmark & & \checkmark & \checkmark  &    \\ 
& prosocial & & \checkmark & \checkmark &\checkmark & \checkmark & \checkmark & \checkmark \\
  \hline
Authority & affect & \checkmark & \checkmark& \checkmark & \checkmark& \checkmark & \checkmark& \checkmark \\ 
&  religion & \checkmark& \checkmark & \checkmark & &\checkmark & \checkmark & \\
& Social & \checkmark & \checkmark & \checkmark & \checkmark & \checkmark & \checkmark & \checkmark \\
   \hline
Purity &  family &  \checkmark& \checkmark& \checkmark& & \checkmark & \checkmark& \\ 
&  religion & \checkmark &\checkmark &\checkmark & \checkmark& \checkmark &\checkmark & \checkmark  \\
   \hline
Care & Social &  &\checkmark & \checkmark &  &  \checkmark& \checkmark & \\ 
&  affiliation & \checkmark & \checkmark  &  \checkmark & & \checkmark  &  \checkmark  &  \\
\multicolumn{9}{c}{\vspace{0.2cm}}\\
\multicolumn{2}{c}{\textbf{Empathy}} & & & & & & & \\
\hline
PD  &  affect & \checkmark &  & & &   &   &  \\ 
          &  differ &  & & & \checkmark &  &   &  \\ 
   \hline
EC & we & \checkmark & & \checkmark & \checkmark & \checkmark & & \checkmark \\
& cogproc & \checkmark &  &  & \checkmark&  &  &    \\ 
& pronoun & \checkmark &  &  & \checkmark&  &  &    \\ 
&  differ &  & &  & \checkmark& &   &    \\ 
   \hline
PT & we & \checkmark & \checkmark & \checkmark & \checkmark & \checkmark & & \checkmark \\
& pronoun & \checkmark & &  & \checkmark& \checkmark & &  \\ 
&  cogproc & &  &  &\checkmark & \checkmark & &\\
   \hline
\end{tabular}}

\end{minipage}
\caption{Highlighted correlations between LIWC categories and author attributes (see \autoref{appendix:top-down-analysis} for exact values and additional categories) on original and LLM-generated texts. \checkmark indicates significant correlations. IRI-fantasy is not displayed as it was not correlated with any dictionary categories.}
\label{tab:checkmarks-personality-morality-empathy}
\end{table}

Adopting a theoretically grounded approach, \autoref{tab:checkmarks-personality-morality-empathy} contains a simplified illustration of some important associations between lexical categories and various dimensions of author attributes in authors' original and LLM-generated texts (see \autoref{appendix:top-down-analysis} for the exact correlation coefficients). 
Focusing on the authors' original texts, we successfully replicated associations that have been previously established across different author attributes, e.g.,  males used a significantly smaller number of social and anxiety-related words than females \citep{ishikawa2015gender}, higher EXT was significantly associated with greater use of positive emotions and social words \citep{chen2020meta}, and higher Loyalty was associated with greater usage of family-related words \citep{LiTomasello}.

Switching to the LLM-generated texts, we observed that few associations were preserved across all utilized LLMs and prompts, such as the association between NEU and negative emotion words, Purity and religion-related words, or number of article words and gender. 
We also noticed occurrences where LLMs washed away linguistic markers of authors' characteristics initially present in the original texts, such as the association between OPN and BigWords, Fairness and religion-related words, and gender and anxiety-related words.
Surprisingly, in a few occurrences, expected associations that were not observed in the original texts emerged in LLM-generated texts, such as the association between age and first-person plural words, NEU and affect-related words, and Loyalty and prosocial words. 
Overall, our results demonstrated that LLM involvement in writing can preserve some theoretically grounded associations while removing others, and in some cases, even add previously established associations that were not observed in the original text. 
For the complete list of correlations, refer to \autoref{appendix:top-down-analysis}. 



\begin{wrapfigure}{r}{0.57\textwidth}
{
  \centering
  \includegraphics[width=0.5\columnwidth]{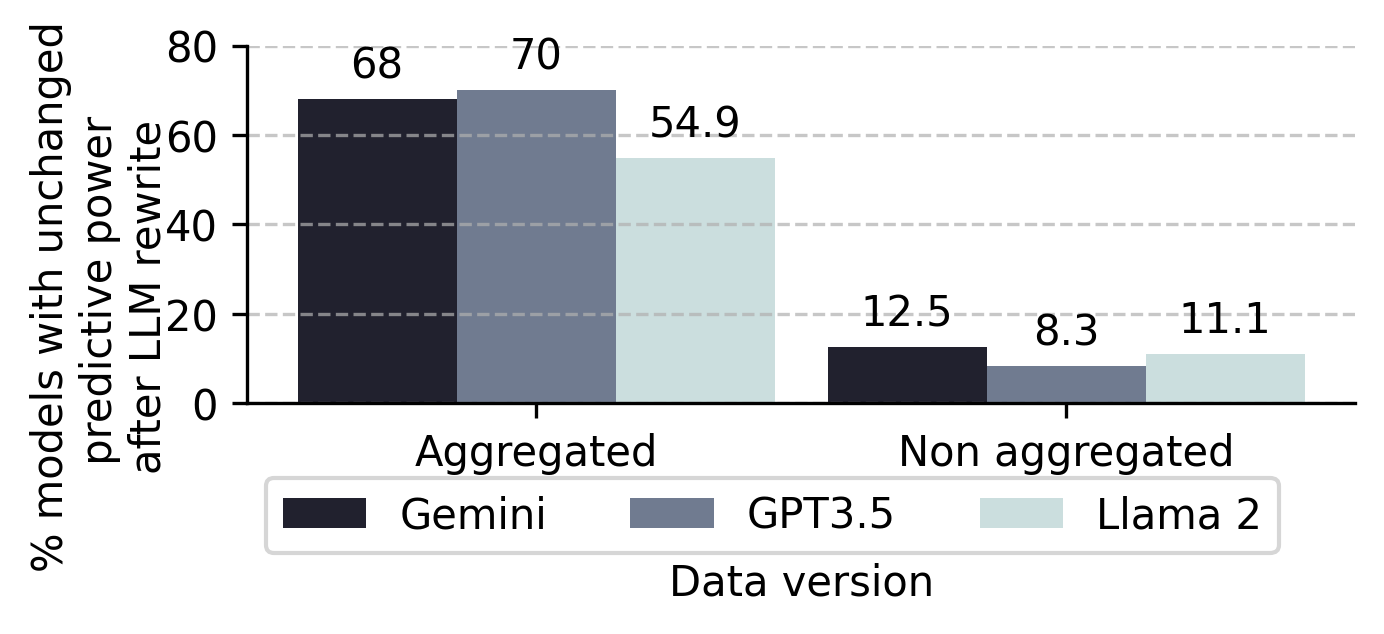}
  \caption{The ratio of unchanged predictive powers for two versions of the Empathetic Conversations dataset, one with aggregated essays per author and one containing all essays from the same author as individual observations.}
  \label{fig:effect-of-data-size}
}
\end{wrapfigure}

\subsection{Sensitivity to Data Size}

The reliability of inferences about authors' personal traits can depend on the amount of text available on authors. We ask if LLMs' impact on the predictive power of linguistic cues is sensitive to the amount of available data. Since the Empathetic Conversations dataset associated with dispositional empathy contains multiple reaction essays from the same authors, we compared the LLMs' influence on both the version of the dataset with essays grouped by authors and the version that treats each essay as an individual observation. \autoref{fig:effect-of-data-size} demonstrates the ratio of classifiers with unchanged predictive power on dispositional empathy for both data versions after LLM rewrites. 
We found that classifiers trained on aggregated texts of authors were less affected by LLMs, as indicated by the ratio of classifiers with unchanged performance (McNemar's $\chi^2 = 91.0, p < .001$, Odds ratio (OR) $ = 0.01$ [large effect size]).

\section{Conclusion}
Previous research links individuals' linguistic patterns with their personal traits, but it's unclear if these links remain unchanged when LLMs are involved in the writing process. Adopting a data-driven approach, our results suggested that LLMs' involvement in writing can slightly reduce the predictive power of linguistic patterns over personal traits, and this significant decline happens infrequently. Additionally, we found that the context in which LLMs are used, the choice of LLM, and the adopted prompt can amplify or rectify the impact of LLMs on the predictive power of linguistic patterns over personal traits. Further fine-grained analysis on lexical-based, theoretically established, linguistic markers of personal traits indicated that LLMs wash away certain linguistic markers, such as the link between fairness and religion-related words or extraversion and social words, while retaining others, such as neuroticism and negative emotion words. Our findings carry significant implications for research on linguistic markers of personal traits in the age of LLMs as they render certain linguistic indicators unreliable, which might prompt reconsideration of utilized methodologies in this area of research.

\section{Limitations}
\label{appendix:limitations}
Although we have tried to cover various contexts, personal traits, LLMs, and prompts, future work could improve the generalizability of this research by including a wider variety of datasets and contexts for each construct of interest, as well as expanding these analyses to additional personal traits. This could help capture the variability of information about personal traits in individuals' linguistic patterns across different contexts, such as Facebook posts, school homework, or formal writing. Future work could also investigate the particular reasons for LLMs' differential effects across various demographic and psychological constructs. 
Moreover, individuals from diverse cultures or backgrounds may exhibit varying degrees of expressiveness in their language use, and different psychological constructs may be more or less apparent in the language of individuals from different cultures, potentially resulting in variability in the linguistic markers' predictive power. Future work should consider the cultural background of the authors, to reduce unaccounted variability and paint a clearer picture of LLMs' impact on linguistic patterns that are predictive of individuals' personal traits. Although we utilized semantic similarity measurements to ensure that LLMs do not change the meaning of authors' original texts, the interplay between different types of similarity between the original and LLM-generated texts, such as semantic similarity and lexical similarity, can provide valuable insights into the overall impact of LLMs on individuals' linguistic patterns, especially in cases where LLMs might alter both meaning and style in writings. 

\section{Ethics Statement}
\label{sec:ethical-statement}
As our research pertains to the identification of private information about authors, we recognize the ethical concerns related to potential misuse, particularly in determining private attributes of authors without their consent. However, the purpose of this study is to enhance our understanding of the impacts of LLMs on this sensitive field of study, and further to contribute to the privacy of the users of these LLMs. It is also noteworthy that the personality, empathy and morality data used in this study were completely anonymized with proper elicitation of consent from authors upon gathering their personal information in the respective studies that gathered this data. The other data source we used for demographic attributes is publicly available. 

\section{Acknowledgements}
This research was supported, in part, by the Army Research Laboratory under contract W911NF-23-2-0183 and by DARPA INCAS HR001121C0165. The views and conclusions contained herein are those of the authors and should not be interpreted as necessarily representing the official policies, either expressed or implied, of DARPA or the U.S. Government. The U.S. Government is authorized to reproduce and distribute reprints for governmental purposes notwithstanding any copyright annotation therein. We also thank James W. Pennebaker for their support and invaluable insights throughout the research process.

\bibliography{colm2024_conference}
\bibliographystyle{colm2024_conference}
\newpage
\appendix

\section{LIWC and Construct-specific Lexicons}
\label{subsec:liwc-categories}

In this section, we present the details of the word categories that were used from LIWC or other construct-specific lexicons in \autoref{do-llms-change-the-associations}. 

\subsection{LIWC \citep{pennebaker2001linguistic}}
The LIWC dictionary is a collection of words categorized into different psychological and linguistic dimensions. It is the backbone of the LIWC software that analyzes text by counting occurrences of words in specific categories, offering insights into the involved emotions, cognition, social interactions, and other linguistic styles. The LIWC categories that we used in this study are as follows:

\begin{itemize}

    \item  Affect: The Affect category is a measure of the emotional tone as well as emotion content of text such as 'happy', 'hate', 'love', and 'wrong'.
    \item  emotion: Emotion words represent a broad category of  emotional expressions referring to specific emotional states, such as 'good', 'love', 'happy', 'hope'.
        
    \item  emo\_neg: Negative emotion words express negative affective states, including terms like 'bad', 'hate', 'hurt', 'tired'.
    
    \item  emo\_pos: Positive emotion words convey positive emotions, including terms like 'happy', 'excited', and 'grateful'.
    
    \item  emo\_anger: Anger words express anger or frustration, including terms like 'hate', 'mad', and 'angry'.
    
     \item  emo\_sad: Emotionally sad words convey feelings of sadness or melancholy, including terms like ':(', 'sad', and 'disappoint'.
     
     \item  swear: Swear words consist of profane or vulgar language used to express strong emotions or taboo subjects, such as 'damn', 'f***', and 'sh*t'.
    
    \item  affiliation: Affiliation words indicate a sense of belonging or connection to others, including terms like 'friend', 'community', and 'support'.
    
    \item socrefs: Social referents encompass words that relate to social interactions, like 'we', 'you', or 'he', as well as words related to family and friends (e.g., 'parent', 'mother', 'girlfriend').
    
    \item  family: Family words relate to familial relationships or members, such as 'mother', 'father', 'sibling', and 'cousin'.

    \item  friend: Friend words relate to friendship or friendly interactions, including terms like 'friend', 'buddy', 'dude', and 'girlfriend'.
        
    \item  pronoun: Pronouns are words used to replace nouns indicating individuals or groups, such as 'I', 'you', 'he', 'she', 'we', and 'they'.

    \item we: This category refers to words indicating group membership or inclusion, such as 'we', 'us', and 'our'.
    
    \item  you: Second-person pronouns, such as 'you' and 'your'.
    
    \item  shehe: Third-person pronouns refer to individuals such as 'he', 'she', 'him', and 'her.'
        
    \item i: First-person singular pronouns refer to the speaker or author, such as 'I', 'me', 'myself', and 'mine'.

    \item  differ: Differentiation words express contrast or distinction between entities or ideas, including terms like 'but', 'not', 'or', and 'if'.

     \item  tentat: Tentative words express uncertainty or hesitation, including terms like 'maybe', 'if', and 'something'.

     \item  cogproc: Cognitive processing words indicate intellectual or cognitive engagement, including terms like 'think', 'understand', 'analyze', and 'consider'.
        
    \item  prosocial: Prosocial words denote behaviors or attitudes that benefit others or society, such as 'help', 'care', 'thank', and 'please'.
        
    \item  BigWords: Big words are words with complex structures (7 letters or longer) that may indicate intellectual complexity or formality, such as 'procrastination', 'circumstantial', and 'phenomenon'.
    
    \item  Drives: Drive words refer to motivational or goal-oriented language, encompassing affiliation (e.g., 'we', 'our', 'us', 'help'), achievement (e.g., 'work', 'better', 'best'), and power (e.g., 'own', 'order', 'allow').
    
    \item  Social: The Social category stands for social processes, and encompasses words pertain to various types of social behavior ('love', 'care', 'please', 'good morning', 'attack', 'deserve', 'judge') and social referents as defined above.

    \item achieve: Achievement words denote actions or concepts related to accomplishment or success, such as 'work', 'bonus', 'beat', and 'overcome'.

    \item inhib: The inhibition category refers to words related to restraint, suppression, and inhibition, such as 'block' and 'constrain'.

     \item  religion: Religion words pertain to religious concepts, practices, or institutions, such as 'God', 'hell', and 'church'.
    
\end{itemize}

\subsection{NRC Emotion Lexicon \citep{Mohammad13,mohammad-turney-2010-emotions}}

\begin{itemize}
    \item nrc.positive: Words that have a positive sentiment, such as 'acceptable', 'boon', and 'civil'.

    \item nrc.negative: Words that have a negative sentiment, such as 'aberrant', 'abort', and 'begging'.

    \item nrc.anger: Words that relate to the emotion of anger, such as 'arguments', 'confront', and 'friction'.

    \item nrc.disgust: Words that relate to the emotion of disgust, such as 'barf', 'decompose', and 'gut'.

    \item nrc.sadness: Words that relate to the emotion of sadness, such as 'blue', 'cloudy', and 'emptiness'.

    \item nrc.anticipation: Words that relate to the emotion of anticipation, such as 'accelerate', 'announcement', and 'approaching'.

    \item nrc.trust: Words that relate to the emotion of trust, such as 'adhering', 'advice', and 'collaborator'.

    \item nrc.joy: Words that relate to the emotion of joy, such as 'whimsical', 'beach', and 'doll'.
\end{itemize}

\subsection{Distress and Empathy Lexicon \citep{sedoc2019learning}}

\begin{itemize}
    \item empathy.low: Low-empathy words from the empathy lexicon (based on a median split from lexicon weights), such as 'joke', 'bizarre', and 'stupidest'.

    \item empathy.high: High-empathy words from the empathy lexicon (based on a median split from lexicon weights), such as 'healing', 'grieve', and 'heartbreaking'. This category did not significantly correlate with any dimension of dispositional empathy.

    \item distress.low: Low-distress words from the distress lexicon (based on a median split from lexicon weights), such as 'dunno', 'guessing', and 'anyhow'.

    \item distress.high: High-distress words from the distress lexicon (based on a median split from lexicon weights), such as 'inhumane', 'dehumanizes', and 'mistreating'. This category did not significantly correlate with any dimension of dispositional empathy.
\end{itemize}

\subsection{Moral Foundations Dictionary 2 (MFD2; \citealp{frimer2019moral})}

\begin{itemize}
    \item mfd.authority.virtue: Words related to the "virtue" dimension of the moral foundation of Authority, such as 'respect', 'obey', and 'honor'.

    \item mfd.authority.vice: Words related to the "vice" dimension of the moral foundation of Authority, such as 'disrespect', 'disobey', and 'chaos'.

    \item mfd.care.virtue: Words related to the "virtue" dimension of the moral foundation of Care, such as 'compassion', 'generosity', and 'pity'.

    \item mfd.care.vice: Words related to the "vice" dimension of the moral foundation of Care, such as 'harm', 'threatens', and 'injured'. This category did not significantly correlate with the Care foundation.

    \item mfd.purity.virtue: Words related to the "virtue" dimension of the moral foundation of Purity, such as 'sacred', 'wholesome', and 'divine'.

    \item mfd.purity.vice: Words related to the "vice" dimension of the moral foundation of Purity, such as 'sin', 'defiled', and 'contaminate'.

    \item mfd.loyalty.virtue: Words related to the "virtue" dimension of the moral foundation of Loyalty, such as 'loyalty', 'allegiance', and 'follower'. This category did not significantly correlate with the Loyalty foundation.

    \item mfd.loyalty.vice: Words related to the "vice" dimension of the moral foundation of Loyalty, such as 'disloyal', 'treason', and 'enemy'. This category did not significantly correlate with the Loyalty foundation.

    \item mfd.fairness.virtue: Words related to the "virtue" dimension of the moral foundation of Fairness, such as 'fairness', 'justice', and 'equality'. This category did not significantly correlate with the Fairness foundation.

    \item mfd.fairness.vice: Words related to the "vice" dimension of the moral foundation of Fairness, such as 'cheat', 'unjust', and 'unequal'. This category did not significantly correlate with the Fairness foundation.
\end{itemize}

\section{Semantic Similarities}

\begin{table*}[ht]
\centering
\begin{tabular}{llrrr}

Dataset & LLM & Median & Mean & SD\\
\hline
 & Gemini & 0.955 & 0.941 & 0.046\\

Essays & GPT3.5 & 0.978 & 0.976 & 0.015\\

& Llama 2 & 0.951 & 0.944 & 0.036\\\hline

 & Gemini & 0.962 & 0.955 & 0.028\\

YourMorals & GPT3.5 & 0.963 & 0.956 & 0.027\\

& Llama 2 & 0.927 & 0.921 & 0.037\\\hline

 & Gemini & 0.971 & 0.967 & 0.022\\

Congress & GPT3.5 & 0.977 & 0.976 & 0.010\\

 & Llama 2 & 0.969 & 0.965 & 0.020\\
\hline
 & Gemini & 0.959 & 0.955 & 0.026\\

Empathetic Conversations & GPT3.5 & 0.966 & 0.964 & 0.014\\

& Llama 2 & 0.934 & 0.928 & 0.031\\
\hline
\end{tabular}
\caption{Descriptive statistics for semantic similarity between the original and LLM-generated texts across different datasets and LLMs.}
\label{tab:similarity_descriptive}
\end{table*}

The descriptive statistics for the semantic similarity between original and LLM-generated texts across different datasets and LLMs is demonstrated in \autoref{tab:similarity_descriptive}.


\begin{table*}[ht]
\centering
\begin{tabular}{lllrr}
Dataset & LLM 1 & LLM 2 & z & Adjusted p-value \\\hline
 & Gemini & GPT3.5 & 40.610 & 0.000\\
Essays & Gemini & Llama 2 & -5.614 & 0.000\\
 & GPT3.5 & Llama 2 & -47.728 & 0.000\\
\hline
 & Gemini & GPT3.5 & 0.123 & 0.902 \\
YourMorals & Gemini & Llama 2 & -41.870 & 0.000\\
 & GPT3.5 & Llama 2 & -41.955 & 0.000\\
\hline
 & Gemini & GPT3.5 & 8.915 & 0.000\\
Congress & Gemini & Llama 2 & -4.506 & 0.000\\
 & GPT3.5 & Llama 2 & -13.421 & 0.000\\
\hline
 & Gemini & GPT3.5 & 2.637 & 0.008 \\
Empathetic Conversations & Gemini & Llama 2 & -6.883 & 0.000\\
& GPT3.5 & Llama 2 & -9.597 & 0.000\\
\hline
\end{tabular}
\caption{Pairwise comparisons between different LLMs, based on the semantic similarity computed between the original and LLM-generated texts in each dataset, with Dunn's test and Benjamini-Hochberg adjustments for multiple comparisons.}
\label{tab:pairwise_sim}
\end{table*}

\begin{table*}[h]
\centering
\begin{tabular}{llcc}
Dataset1 & Dataset2 & z & Adjusted p-value\\
\hline
Essays & yourMorals & -25.314 & \textbf{0.000}\\

Essays & Congress & 21.479 & \textbf{0.000}\\

Essays & Empathetic. & -7.138 & \textbf{0.000}\\

yourMorals & Congress & 39.110 & \textbf{0.000} \\

yourMorals & Empathetic. & -1.842 & 0.065\\

Congress & Empathetic. & -13.589 & \textbf{0.000}\\
\hline
\end{tabular}
\caption{Pairwise comparisons between different datasets, based on the semantic similarity computed between the original and LLM-generated texts in each dataset, with Dunn's test and Benjamini-Hochberg adjustments for multiple comparisons.}
\label{tab:pairwise_sim_datasets}
\end{table*}

Analysis of the pairwise comparison between different LLMs and different datasets, based on the semantic similarity between original and LLM-generated texts, are demonstrated in \autoref{tab:pairwise_sim} and \autoref{tab:pairwise_sim_datasets}, respectively.

\section{Predictive Powers}
The difference between classifiers' $F_1$ scores on the original text and LLM-generated texts, on average, across different psychological and demographic attributes of authors are demonstrated in \autoref{tab:ttest_res}.

\begin{table*}[h!]
\centering
\resizebox{\textwidth}{!}{
\begin{tabular}{lccccrrrrrrrl}
  & $\text{Mean $F_1$}_{\text{original}}$ & $\text{Mean $F_1$}_{\text{LLM}}$ & Mean difference & CI & SE & t & df & p & \emph{d} & $\text{N}_{\text{original}}$ & $\text{N}_{\text{LLM}}$ & magnitude\\
\midrule
Age group & 0.351 & 0.260 & 0.091 & [0.079, 0.102] & 0.006 & 15.397 & 804 & 0 & 0.543 & 805 & 805 & moderate\\

Empathy & 0.657 & 0.603 & 0.054 & [0.050, 0.057] & 0.002 & 32.158 & 6556 & 0 & 0.397 & 6557 & 6557 & small\\

Personality & 0.658 & 0.602 & 0.056 & [0.052, 0.060] & 0.002 & 30.046 & 7652 & 0 & 0.343 & 7653 & 7653 & small\\

Gender & 0.694 & 0.623 & 0.072 & [0.063, 0.080] & 0.004 & 15.965 & 1371 & 0 & 0.431 & 1372 & 1372 & small\\

Morality & 0.664 & 0.591 & 0.062 & [0.057, 0.066] & 0.002 & 25.601 & 5195 & 0 & 0.355 & 5196 & 5196 & small\\

Affiliation & 0.640 & 0.578 & 0.073 & [0.064, 0.082] & 0.005 & 15.691 & 1313 & 0 & 0.433 & 1314 & 1314 & small\\
\bottomrule
\end{tabular}}
\caption{Paired $t$-tests for testing the difference in predictive powers ($F_1$) of classifiers on the original and LLM-generated texts, for each personal trait and Cohen's \emph{d} effect sizes for the magnitude of these differences ($|d|<0.2$: negligible, $|d|<0.5$: small, $|d|<0.8$: moderate; \citealp{cohen}).}
\label{tab:ttest_res}
\end{table*}

\begin{figure*}[h!]
    \centering
    \includegraphics[width=\textwidth]{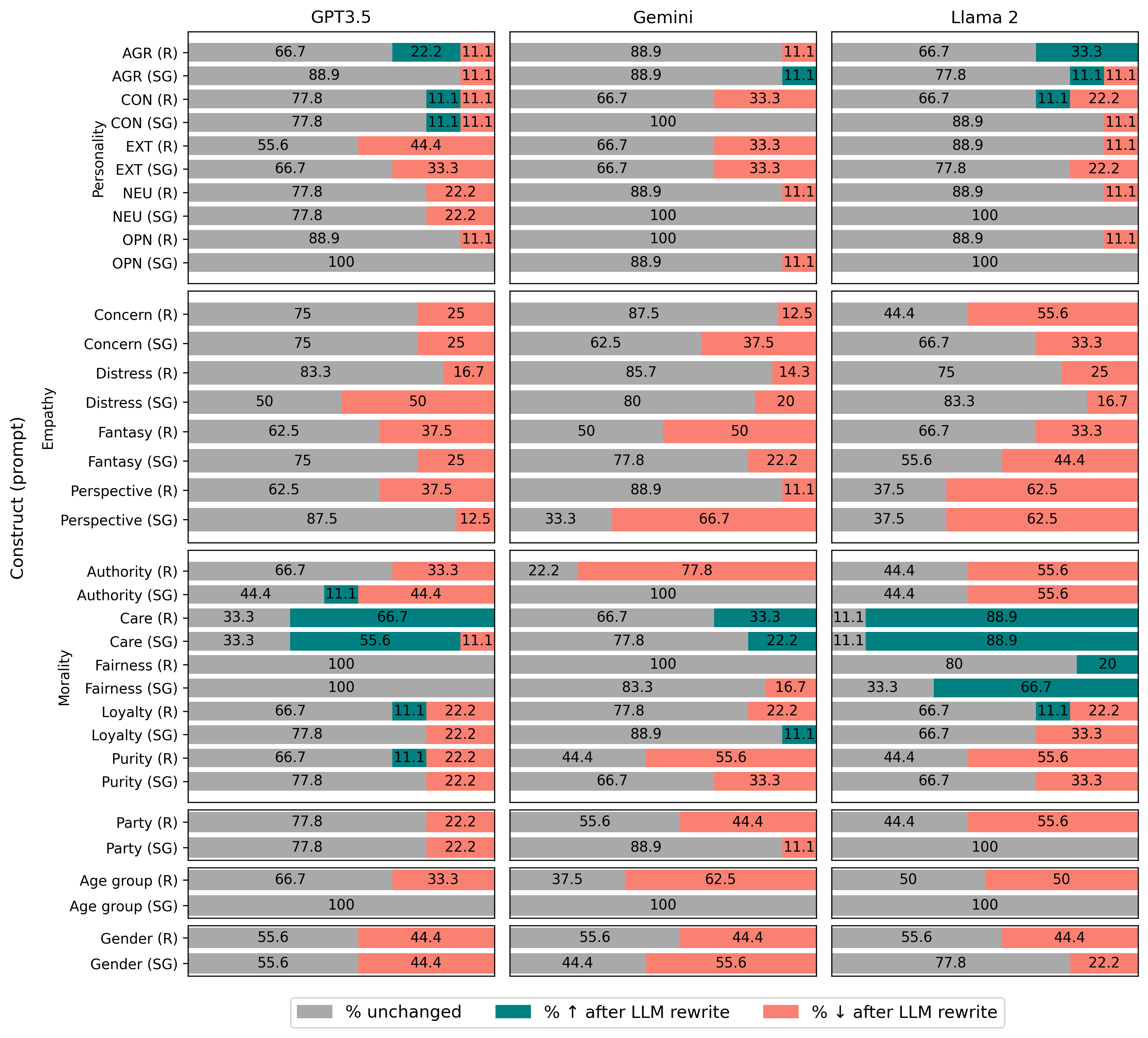}
    \caption{The ratio of models with unchanged, reduced, and enhanced predictive powers after LLM rewrite across granular dimensions of author attributes, using three LLMs and two prompts: \textbf{R}ephrase and \textbf{S}yntax\_\textbf{G}rammar.}
    \label{fig:results-for-differences-for-all-labels}
\end{figure*}

The impact of LLMs on the predictive power of linguistic patterns over authors' personal traits, focusing on different dimensions of investigated psychological and demographic attributes, is shown in \autoref{fig:results-for-differences-for-all-labels}.

\clearpage
\section{Top-down Analyses}
\label{appendix:top-down-analysis}
\begin{table*}[h]
\centering
\resizebox{\textwidth}{!}{%
\begin{tabular}{llr|rrr|rrr}
&& \multicolumn{1}{c}{ }& \multicolumn{3}{c}{Rephrase} & \multicolumn{3}{c}{Syntax-Grammar} \\
 \cmidrule(lr){4-6}\cmidrule(l){7-9}
 && \multicolumn{1}{c}{Original}& \multicolumn{1}{c}{Gemini}& \multicolumn{1}{c}{GPT3.5} &\multicolumn{1}{c}{Llama 2} & \multicolumn{1}{c}{Gemini}& 
   \multicolumn{1}{c}{GPT3.5} & \multicolumn{1}{c}{Llama 2} \\
  \hline
OPN & i & \bf{-0.13} & -0.07 & \bf{-0.11} &  -0.07 & \bf{-0.14} &  \bf{-0.12}  & \bf{-0.09}  \\ 
&  affiliation & -0.06  & \bf{-0.10}  & \bf{-0.07}  & -0.06 &  -0.06 &  \bf{-0.08} & -0.06 \\ 
&  achieve & -0.05 & \bf{-0.08} & \bf{-0.08} & -0.03 & -0.04 & \bf{-0.08} & -0.07 \\ 
&  drives & -0.04 & \bf{-0.10} & \bf{-0.07} & -0.04 & -0.05 & \bf{-0.07}  & -0.05  \\ 
&  swear & \bf{0.09} & 0.04 &  \bf{0.09}  & 0.06 & \bf{0.08} &  0.07 & \bf{0.09} \\ 
&  BigWords & \bf{0.09} & 0.04 & 0.00 & 0.04& 0.07 & -0.01 & 0.03 \\
& nrc.anticipation& -0.04& -0.04& \bf{-0.08}& -0.01& -0.04&  \bf{-0.08}& -0.04\\
& nrc.disgust& \bf{0.09}& 0.05& 0.03& 0.03& 0.06& 0.03& 0.07\\
& nrc.trust& -0.05& -0.03& \bf{-0.08} & 0.00& -0.05& -0.07&  -0.03\\ 
\hline
CON & emo\_anger & \bf{-0.12}  & \bf{-0.12}  & \bf{-0.11} & -0.04 & -0.06 & \bf{-0.12} & \bf{-0.08}  \\ 
&  swear & \bf{-0.11} & -0.01 & \bf{-0.07}  & -0.03 & -0.05 &  \bf{-0.09}  & \bf{-0.08} \\
& nrc.disgust& \bf{-0.08}& -0.07& \bf{-0.08}& -0.03& 0.00& \bf{-0.07}& -0.04\\
& nrc.sadness& \bf{-0.07} & -0.03& -0.07& -0.04& 0.02& -0.05&  -0.03\\ 
\hline
EXT & affiliation & \bf{0.11} & \bf{0.09} & \bf{0.11} & \bf{0.09} & \bf{0.10} & \bf{0.11} & \bf{0.13} \\ 
& drives & \bf{0.09} & 0.07  & \bf{0.07} & \bf{0.08}  & \bf{0.08} & \bf{0.09}  & \bf{0.10}  \\ 
& social & \bf{0.07}  & 0.01 & \bf{0.08}  & 0.06 & 0.04  & 0.06 & 0.07 \\ 
&  affect & 0.04 & 0.06  & \bf{0.08}  & \bf{0.08}  & 0.07& 0.07  & 0.06  \\ 
&  emotion & 0.06 & 0.04 & \bf{0.07}  & 0.06  & 0.05 & \bf{0.07} & \bf{0.07}  \\ 
&  emo\_pos & \bf{0.10}  & \bf{0.09}  & \bf{0.12}  & \bf{0.10} &   0.05  & \bf{0.11} & \bf{0.09}  \\
& nrc.positive& 0.03& 0.03& 0.06& \bf{0.08}& 0.04& 0.07& 0.05\\
& nrc.joy& \bf{0.08}& 0.07& \bf{0.09}& \bf{0.11}& 0.07& \bf{0.09}& \bf{0.08}\\
& nrc.trust& 0.02& 0.03& 0.03& \bf{0.08}& 0.02& 0.06& 0.02\\ 
\hline
AGR &affiliation & \bf{0.08}  & 0.07  & \bf{0.09}  & \bf{0.07} &  \bf{0.08}  & \bf{0.08} & \bf{0.07}  \\ 
&  emo\_anger & \bf{-0.09}  & \bf{-0.08}  & \bf{-0.11} & \bf{-0.08} & \bf{-0.10} & \bf{-0.08}  & \bf{-0.10}  \\ 
&  swear & \bf{-0.12}  & -0.03 & -0.06 & -0.03  & -0.08  & \bf{-0.08}  & \bf{-0.08}  \\
 & nrc.negative& \bf{-.010} & -0.04& -0.05& -0.06& \bf{-0.08} & -0.07& -0.07\\
 & nrc.anticipation& 0.07& 0.04& \bf{0.08} & 0.03& 0.04&  0.06& \bf{0.07}\\
 & nrc.disgust& \bf{-0.08} & -0.06& -0.05& -0.05& -0.05& -0.07&  -0.05\\ 
   \hline
NEU & i & \bf{0.15} & -0.02  & \bf{0.16} & \bf{0.15}  & \bf{0.15} &  \bf{0.14}  & \bf{0.12} \\ 
 & pronoun & \bf{0.12}& 0.01 & \bf{0.11}  & \bf{0.09} & \bf{0.10}  & \bf{0.09}  & \bf{0.10}  \\ 
&  affect & 0.06  & \bf{0.11}& \bf{0.07} & 0.06 & 0.07& \bf{0.08}  & 0.03 \\ 
 & emotion & \bf{0.10} & \bf{0.14}  & \bf{0.12}  & \bf{0.11} & \bf{0.10}& \bf{0.12}  & 0.07 \\ 
&  emo\_neg & \bf{0.19}  & \bf{0.18} & \bf{0.20}  & \bf{0.21} &  \bf{0.15} & \bf{0.21}  & \bf{0.18}  \\ 
 & emo\_anger & \bf{0.07}  & 0.06 & 0.07  & 0.05 & 0.06  & 0.07 & 0.06\\ 
&  emo\_sad & \bf{0.10} & \bf{0.10}  & \bf{0.10} & \bf{0.10} &  \bf{0.08}  & \bf{0.10} & 0.05 \\
 & nrc.positive & -0.07& -0.02& -0.06& \bf{-0.08} & -0.04& -0.04&  \bf{-0.09}\\
 & nrc.negative & 0.05& \bf{0.11} & \bf{0.09} & \bf{0.15}& 0.05&  \bf{0.08} & \bf{0.07} \\
 & nrc.anger & \bf{0.08}& \bf{0.14}& \bf{0.12} & \bf{0.15}&  0.07& \bf{0.10} & \bf{0.08} \\
 & nrc.disgust& 0.07& \bf{0.08}& 0.07& 0.07& 0.02& \bf{0.08} & 0.05\\\hline
\end{tabular}}
\caption{Pearson correlations between LIWC and NRC dictionary word frequencies, and the Big Five personality dimensions, on the authors' original texts and LLM-generated texts, across different utilized LLMs and prompts. Significant correlations are \textbf{boldfaced}.} 
\label{tab:complete-table-personality}
\end{table*}

For personality, using the original texts, we replicated previously known associations in the literature \citep[e.g.,][]{baddeley_singer,hirsh2009personality,tackman}. Namely, higher OPN was significantly associated with more complex language usage (BigWords) and swear words, higher CON was significantly associated with fewer swear and negative emotion words, higher EXT was significantly associated with greater use of positive emotion and social words, higher AGR was significantly associated with fewer swear and negative emotion words, and higher NEU was significantly associated with greater use of negative emotion words and pronouns.
In general, expected associations appear to mostly have been preserved with the involvement of GPT3.5 and Gemini, while Llama 2 appears to have preserved fewer of these associations. Certain associations present in the original texts were retained regardless of the LLM involved or prompt used (e.g., EXT and affiliation-related words; AGR and anger emotion words and NEU and negative emotion words). Other associations disappeared, regardless of the specific LLM or prompt (OPN and BigWords, and NEU and anger emotion words). 
Overall, these results suggest that the fine-grained lexical cues that allow for personality detection using theoretically grounded approaches might not be reliable when LLMs are involved in the writing process, and might be somewhat dependent on which LLM users choose to utilize and the personality dimension of interest. 


\begin{table*}[h!]
\centering
\resizebox{\textwidth}{!}{
\begin{tabular}{llr|rrr|rrr}
&& \multicolumn{1}{c}{ }& \multicolumn{3}{c}{Rephrase} & \multicolumn{3}{c}{Syntax-Grammar} \\
 \cmidrule(lr){4-6}\cmidrule(l){7-9}
& & \multicolumn{1}{c}{Original}& \multicolumn{1}{c}{Gemini}& \multicolumn{1}{c}{GPT3.5} &\multicolumn{1}{c}{Llama 2} & \multicolumn{1}{c}{Gemini}& 
\multicolumn{1}{c}{GPT3.5} & \multicolumn{1}{c}{Llama 2} \\
\hline
Fairness & pronoun & -0.05& -0.03& -0.01 & -0.02 & \bf{-0.06} & -0.02 & -0.02 \\ 
&  religion & \bf{-0.06}& -0.03& -0.05 & -0.01 & -0.05 & -0.06 & -0.02 \\ 
  \hline
Loyalty& i & 0.03& 0.03& \bf{0.06} & 0.10 & 0.04 & 0.05 & 0.11 \\ 
&  you & \bf{0.08}& \bf{0.10}& \bf{0.09} & \bf{0.15} & \bf{0.09} & \bf{0.10} & \bf{0.12} \\ 
&  pronoun & 0.04& 0.05& \bf{0.06} & \bf{0.14} & 0.05 & \bf{0.08} & 0.09 \\ 
&  affiliation & \bf{0.06}& \bf{0.10}& \bf{0.08} & \bf{0.11} & \bf{0.07} & 0.06 & 0.07 \\ 
&  social & \bf{0.08}& \bf{0.12} & \bf{0.10} & \bf{0.17} & \bf{0.10} & \bf{0.11} & \bf{0.12} \\ 
&  prosocial & 0.05& \bf{0.07} & \bf{0.06} & \bf{0.15} & \bf{0.06} & \bf{0.07} & \bf{0.11} \\ 
&  religion & \bf{0.12}& \bf{0.12} & \bf{0.13} & 0.09 & \bf{0.12} & \bf{0.12} & \bf{0.11} \\ 
&  affect & \bf{0.10}& \bf{0.10} & \bf{0.13} & \bf{0.21} & \bf{0.12} & \bf{0.11} & \bf{0.11} \\ 
&  socrefs & \bf{0.10}& \bf{0.13} & \bf{0.11} & \bf{0.16} & \bf{0.11} & \bf{0.10} & 0.09 \\ 
&  family & \bf{0.11}& \bf{0.12} & \bf{0.12} & 0.09 & \bf{0.11} & \bf{0.11} & 0.10 \\ 
&  friend & \bf{0.07}& \bf{0.06} & 0.06 & 0.07 & \bf{0.06} & 0.06 & 0.06 \\ 
   \hline
Authority &i & \bf{0.07}& \bf{0.07} & \bf{0.09} & \bf{0.12} & \bf{0.08} & \bf{0.07} & \bf{0.11} \\ 
&  you & \bf{0.11}& \bf{0.11} & \bf{0.09} & \bf{0.19} & \bf{0.11} & \bf{0.09} & \bf{0.14} \\ 
&  pronoun & \bf{0.13}& \bf{0.11} & \bf{0.11} & \bf{0.18} & \bf{0.13} & \bf{0.12} & \bf{0.12} \\ 
&  affiliation & \bf{0.12}& \bf{0.11} & \bf{0.11} & \bf{0.14} & \bf{0.11} & \bf{0.09} & 0.10 \\ 
&  Social & \bf{0.16}& \bf{0.14} & \bf{0.12} & \bf{0.17} & \bf{0.16} & \bf{0.12} & \bf{0.14} \\ 
&  prosocial & \bf{0.06}& \bf{0.07} & 0.05 & \bf{0.13} & \bf{0.07} & 0.06 & \bf{0.12} \\ 
&  religion & \bf{0.16}& \bf{0.15} & \bf{0.15} & 0.06 & \bf{0.16} & \bf{0.16} & 0.09 \\ 
&  affect & \bf{0.15}& \bf{0.13} & \bf{0.13} & \bf{0.25} & \bf{0.16} & \bf{0.13} & \bf{0.15} \\ 
&  socrefs & \bf{0.16}& \bf{0.15} & \bf{0.14} & \bf{0.19} & \bf{0.16} & \bf{0.13} & \bf{0.12} \\ 
&  family & \bf{0.16}& \bf{0.15} & \bf{0.17} & \bf{0.14} & \bf{0.15} & \bf{0.16} & \bf{0.14} \\ 
&  friend & \bf{0.07}& 0.06 & 0.05 & 0.08 & 0.06 & 0.04 & 0.07 \\
 & mfd.authority.virtue& \bf{0.04}& \bf{0.04}& 0.00& 0.01& \bf{0.04}& 0.02& 0.03\\
 & mfd.authority.vice& \bf{0.05}& \bf{0.05}& \bf{0.04}& 0.03& \bf{0.05}& \bf{0.04}& 0.02\\ 
   \hline
Purity & i & \bf{0.06}& 0.04 & \bf{0.06} & \bf{0.11} & 0.05 & \bf{0.06} & 0.11 \\ 
&  you & \bf{0.13}& \bf{0.13} & \bf{0.11} & \bf{0.17} & \bf{0.13} & \bf{0.12} & \bf{0.15} \\ 
&  we & \bf{0.07}& \bf{0.09} & \bf{0.09} & 0.07 & \bf{0.08} & \bf{0.07} & 0.02 \\ 
&  pronoun & \bf{0.13}& \bf{0.11} & \bf{0.13} & \bf{0.17} & \bf{0.13} & \bf{0.14} & \bf{0.11} \\ 
&  affiliation & \bf{0.15}& \bf{0.15} & \bf{0.15} & \bf{0.14} & \bf{0.14} & \bf{0.12} & \bf{0.10} \\ 
&  social & \bf{0.19}& \bf{0.20} & \bf{0.18} & \bf{0.18} & \bf{0.20} & \bf{0.18} & \bf{0.16} \\ 
&  prosocial & \bf{0.06}& \bf{0.07} & 0.05 & \bf{0.12} & \bf{0.08} & \bf{0.06} & \bf{0.13} \\ 
&  religion & \bf{0.21}& \bf{0.21} & \bf{0.21} & \bf{0.11} & \bf{0.21} & \bf{0.22} & \bf{0.16} \\ 
&  affect & \bf{0.12}& \bf{0.12} & \bf{0.13} & \bf{0.20} & \bf{0.15} & \bf{0.11} & \bf{0.12} \\ 
&  socrefs & \bf{0.19}& \bf{0.20} & \bf{0.20} & \bf{0.19} & \bf{0.20} & \bf{0.19} & \bf{0.13} \\ 
&  family & \bf{0.15}& \bf{0.15} & \bf{0.17} & 0.10 & \bf{0.15} & \bf{0.15} & 0.10 \\ 
&  friend & \bf{0.07}& \bf{0.07} & \bf{0.06} & 0.08 & \bf{0.07} & \bf{0.06} & 0.10 \\
 & mfd.purity.virtue& \bf{0.21}& \bf{0.19}& \bf{0.22}& \bf{0.11}& \bf{0.21}& \bf{0.22}& \bf{0.18}\\
 & mfd.purity.vice& \bf{-0.06}& 0.01& 0.02& -0.02& 0.00& -0.02& -0.05\\ 
\hline
Care & we & 0.03& \bf{0.06} & 0.05 & -0.01 & 0.05 & 0.03 & 0.04 \\ 
 & pronoun & 0.03& \bf{0.07} & \bf{0.06} & 0.08 & 0.04 & \bf{0.07} & 0.06 \\ 
 & affiliation & \bf{0.07}& \bf{0.09} & \bf{0.10} & 0.06 & \bf{0.08} & \bf{0.07} & 0.07 \\ 
 & social & 0.05& \bf{0.09} & \bf{0.10} & 0.07 & \bf{0.07} & \bf{0.09} & 0.08 \\ 
 & socrefs & 0.05& \bf{0.08} & \bf{0.09} & 0.09 & \bf{0.06} & \bf{0.08} & 0.07 \\ 
&  family & 0.05& 0.05 & \bf{0.07} & 0.09 & \bf{0.06} & \bf{0.06} & 0.08 \\ 
 & friend & 0.06& \bf{0.06} & \bf{0.07} & 0.06 & 0.05 & 0.04 & 0.05 \\
 & mfd.care.virtue & \bf{0.09}& \bf{0.09}& \bf{0.10}& 0.02& \bf{0.07}& \bf{0.09}& \bf{0.08}\\\hline
\end{tabular}}
\caption{Pearson correlations between relevant dictionary categories and the moral foundations of Fairness, Loyalty, Authority, Purity, and Care before and after LLM (Gemini, GPT3.5, Llama 2) rewrite. Significant correlations are \textbf{boldfaced}.} 
\label{tab:complete-table-morality}
\end{table*}

In line with evolutionary accounts of morality linking the development of moral values to the necessity of cooperation and interdependence \citep{LiTomasello}, we found expected significant associations between MFT dimensions and social and affiliation-related words (e.g., between Loyalty and family-related words), and between foundations such as Purity and Authority and word categories such as religion-related words. We found fewer hypothesized associations in the Fairness and Care foundation. In general, most of the hypothesized associations across all foundations were preserved with LLM involvement and were even amplified for the Care foundation, specifically using Gemini and GPT3.5. 
Nevertheless, some associations present in the original texts were washed away after LLM involvement, primarily with Llama 2 (e.g., between Purity and family-related words, Authority and religion-related words). These results are aligned with our observations in \autoref{do-llms-wash-away-stuff}, underscoring Gemini and GPT3.5 as more preservative of lexical cues predictive of authors' moral values, than Llama 2, and also associations between Care and people's language being enhanced when LLMs are involved in the writing process.

\begin{table*}[ht]
\centering
\resizebox{\textwidth}{!}{%
\begin{tabular}{llr|rrr|rrr}

&& \multicolumn{1}{c}{ }& \multicolumn{3}{c}{Rephrase} & \multicolumn{3}{c}{Syntax-Grammar} \\
 \cmidrule(lr){4-6}\cmidrule(l){7-9}
& & \multicolumn{1}{c}{Original}& \multicolumn{1}{c}{Gemini}& \multicolumn{1}{c}{GPT3.5} &\multicolumn{1}{c}{Llama 2} & \multicolumn{1}{c}{Gemini}& 
\multicolumn{1}{c}{GPT3.5} & \multicolumn{1}{c}{Llama 2} \\
\hline
PD & affect & \bf{0.13} & 0.06  & 0.08 & 0.07 &  0.06  & 0.09  & 0.12  \\ 
&  differ & 0.09  & 0.11 & 0.12  & \bf{0.16}  & 0.10  & 0.09 &  0.08  \\ 
   \hline
EC & we & \bf{-0.22} & -0.13  & \bf{-0.20} & \bf{-0.17} & \bf{-0.21} & -0.11 & \bf{-0.15}  \\ 
&  she,he & \bf{0.14} & 0.08  & 0.11& \bf{0.14} & 0.13  & 0.07 &  0.13  \\ 
&  pronoun & \bf{-0.15} & 0.00 & -0.07 & \bf{-0.13}  & -0.11 &  0.00  & -0.05\\ 
&  emo\_neg & 0.13 & 0.07  & 0.13  & \bf{0.14}  & 0.12  & 0.11 &  0.11  \\ 
&  cogproc & -\bf{0.13}& -0.12 & -0.13 & \bf{-0.14}  & -0.13 & -0.12& -0.07 \\ 
&  tentat & \bf{-0.13} & -0.12  & -0.13 & \bf{-0.16} & \bf{-0.17}  & -0.02 & -0.10  \\ 
&  differ & -0.10 &  -0.07  & -0.08 & \bf{-0.14} & -0.13  & -0.09  & -0.11 \\
 & empathy.low & -0.04& -0.09& -0.04& -0.10& -0.03& -0.02& \bf{-0.14} \\
 & distress.low & -0.09& -0.09& -0.04& -0.11& -0.05& -0.03& \bf{-0.16} \\ 

\hline
PT & we & \bf{-0.22}  & \bf{-0.17} & \bf{-0.19}  & \bf{-0.17} & \bf{-0.22}  & -0.13  & \bf{-0.18}  \\ 
&  pronoun & \bf{-0.22} & -0.06  & -0.14& \bf{-0.20} & \bf{-0.17} & -0.03  & -0.07 \\ 
&  cogproc & -0.12 & -0.14 & -0.13& \bf{-0.18} & \bf{-0.14}  & -0.10  & -0.08\\
 & distress.low &  0.07& -0.05& -0.02& -0.12& 0.05& 0.01&  \bf{-0.14} \\\hline
\end{tabular}}

\caption{Pearson correlations between word frequencies of LIWC categories and personal distress (PD), empathetic concern (EC), and perspective-taking (PT) before and after LLM (Gemini, GPT3.5, Llama 2) rewrites. Significant correlations are \textbf{boldfaced}. The fourth dimension of the IRI, Fantasy, is not displayed as no LIWC or empathy/distress lexicon categories were significantly correlated with this dimension.} 
\label{tab:complete-table-empathy}
\end{table*}

For dimensions of dispositional empathy, we expected and found significant associations with the use of pronouns, emotion words, and words related to cognitive processes (i.e., cogproc), as well as with words from the empathy and distress lexicons. Hypothesized associations were present in the original text for three subdimensions of IRI (e.g., between PD and affect-related words, EC and pronouns, as well as tentative-related words (tentat), and PT and pronouns). The significant negative association of IRI dimensions PT and EC with first-person plural words (we) was the only association that retained its significance across almost all LLM rewrite conditions. An expected association between PT and cogproc-related words, which had not been present in the original text, emerged after involving Gemini and Llama 2. Among different LLMs and prompts, involving Llama 2 with a Rephrase prompt in writing retained the most significant correlations, similar to the trend observed in \autoref{fig:overall-results-for-changes-in-predictive-power} (e.g., EC with cogproc-related words and pronouns), and also created more additional correlations (e.g., between PD and words showing differentiation; differ for short). 
No categories significantly correlated with the FS dimension in the original or in the LLM-generated text.

\begin{table*}[h!]
\centering
\resizebox{\textwidth}{!}{%
\begin{tabular}{lrr|rrrrrr|rrrrrr}
& \multicolumn{2}{c}{} & \multicolumn{6}{c}{Rephrase}& \multicolumn{6}{c}{Syntax-Grammar} \\ 
\cmidrule(lr){4-9} \cmidrule(l){10-15}
& \multicolumn{2}{c}{Original} & \multicolumn{2}{c}{Gemini}& \multicolumn{2}{c}{GPT3.5} & \multicolumn{2}{c}{Llama 2} & \multicolumn{2}{c}{Gemini}& \multicolumn{2}{c}{GPT3.5} & \multicolumn{2}{c}{Llama 2} \\ \hline
& $M_D$ & $M_R$ & $M_D$ & $M_R$ &  $M_D$ & $M_R$ &  $M_D$ & $M_R$ &  $M_D$ & $M_R$ &  $M_D$ & $M_R$ & $M_D$ & $M_R$  \\ \hline
emo\_anx & 0.07 & 0.06 & \bf{0.08} & \bf{0.05} & 0.07 & 0.07 & 0.06 & 0.06 & \bf{0.07} & \bf{0.06} & 0.08 & 0.07 & 0.07 & 0.06 \\
 adverb& \bf{3.41}& \bf{3.71}& 1.98& 2.07& 2.22& 2.35& \bf{2.31}& \bf{2.45}& \bf{3.06}& \bf{3.28}& \bf{2.38}& \bf{2.52}& \bf{2.82}&\bf{3.04}\\
 conj& 5.75& 5.69& \bf{5.48}& \bf{5.25}& \bf{5.37}& \bf{5.15}& 6.11& 5.95& 5.60& 5.51& \bf{5.24}& \bf{5.04}& 5.75&5.64\\
 \bottomrule
\end{tabular}
}
\caption{Average word frequencies (\textbf{$M$}ean) for relevant LIWC categories in two political affiliations (Democrat or Republican) before and after LLM (Gemini, GPT3.5, Llama 2) rewrites. Significant t-tests are \textbf{boldfaced}.} 
\label{tab:correlation_pol}
\end{table*}
\begin{table*}[h!]
\centering
\resizebox{\textwidth}{!}{%
\begin{tabular}{lrr|rrrrrr|rrrrrr}
& \multicolumn{2}{c}{}& \multicolumn{6}{c}{Rephrase}& \multicolumn{6}{c}{Syntax-Grammar}\\ 
\cmidrule(lr){4-9} \cmidrule(l){10-15}
& \multicolumn{2}{c}{Original}& \multicolumn{2}{c}{Gemini}& \multicolumn{2}{c}{GPT3.5}& \multicolumn{2}{c}{Llama 2}& \multicolumn{2}{c}{Gemini}& \multicolumn{2}{c}{GPT3.5}& \multicolumn{2}{c}{Llama 2}\\ 
\hline
& $M_M$& $M_F$& $M_M$& $M_F$& $M_M$& $M_F$& $M_M$& $M_F$& $M_M$& $M_F$& $M_M$& $M_F$& $M_M$& $M_F$\\ 
\hline
article& \bf{8.52}& \bf{8.01}& \bf{8.72}& \bf{8.20}& \bf{8.45}& \bf{7.99}& \bf{8.98}& \bf{8.41}& \bf{8.60}& \bf{8.01}& \bf{8.79}& \bf{8.31}& \bf{8.75}& \bf{8.17}\\
   
social& \bf{10.8}& \bf{11.5}& \bf{11.1}& \bf{11.9}& \bf{11.0}& \bf{11.6}& 12.0& 12.5& \bf{10.9}& \bf{11.7}& \bf{10.4}& \bf{10.9}& \bf{11.3}&\bf{12.0}\\
emo\_anx& \bf{0.06}& \bf{0.07}& \bf{0.05}& \bf{0.08}& 0.06& 0.08& 0.05& 0.07& 0.06& 0.07& 0.07& 0.08& 0.06&0.07\\
\bottomrule
\end{tabular}
}
\caption{Average word frequencies (\textbf{$M$}ean) for relevant LIWC categories in two investigated categories of gender (Male and Female) before and after LLM (Gemini, GPT3.5, Llama 2) rewrites. Significant t-tests are \textbf{boldfaced}.} 
\label{tab:correlation_gender}
\end{table*}
\begin{table*}[ht]
\centering
\resizebox{\textwidth}{!}{%
\begin{tabular}{lr|rrr|rrr}
& \multicolumn{1}{c}{ }& \multicolumn{3}{c}{Rephrase (R)} & \multicolumn{3}{c}{Syntax-Grammar (SG)} \\
\cmidrule(lr){3-5} \cmidrule(l){6-8}
& \multicolumn{1}{c}{Original}& \multicolumn{1}{c}{Gemini}& \multicolumn{1}{c}{GPT3.5} &\multicolumn{1}{c}{Llama 2} & \multicolumn{1}{c}{Gemini}& 
\multicolumn{1}{c}{GPT3.5} & \multicolumn{1}{c}{Llama 2} \\
\hline
we& -0.07& \bf{--0.16}& -0.10& \bf{--0.12}&  -0.06& \bf{--0.12}& -0.07\\
cogproc& -0.06& \bf{--0.13}& \bf{-0.11}& \bf{-0.11}& -0.08& \bf{-0.15}&-0.10\\
\bottomrule
\end{tabular}}

\caption{Pearson correlations between LIWC categories and age before and after LLM (Gemini, GPT3.5, Llama 2) rewrites. Significant correlations are \textbf{boldfaced}.} 
\label{tab:complete-table-age}
\end{table*}

For demographic variables, we found several expected significant differences in the usage of lexical cues across political affiliations (we did not find hypothesized differences across political affiliation in language use related to inhibition and reaction to threats, e.g., negative emotion words; \citealp{okdie}) and genders. Namely, we found that Republicans used a significantly greater number of adverbs than Democrats, and males used a significantly greater number of articles and a significantly smaller number of social and anxiety-related words than females \citep{ishikawa2015gender}. We did not find any hypothesized associations with age.
After the LLM rewrites, some previously expected associations for political affiliation retained significance, while others were washed away. The linguistic cues for gender were generally preserved for article and social word usage, but only Gemini using a Rephrase prompt, preserved the linguistic cues for anxiety-related words. Interestingly, after LLM rewrite, we found that some hypothesized correlations with age appeared; namely, first-person plural words and words related to cognitive processes (e.g., but, not, if, or, know) significantly correlated with age after LLM rewrite in some conditions. In general, it seems as though Gemini and Llama 2 amplified these correlations using a Rephrase prompt, and GPT3.5 amplified these correlations using a Syntax\_Grammar prompt.

\begin{figure}[ht]
\centering
\includegraphics[width=1.0\textwidth]{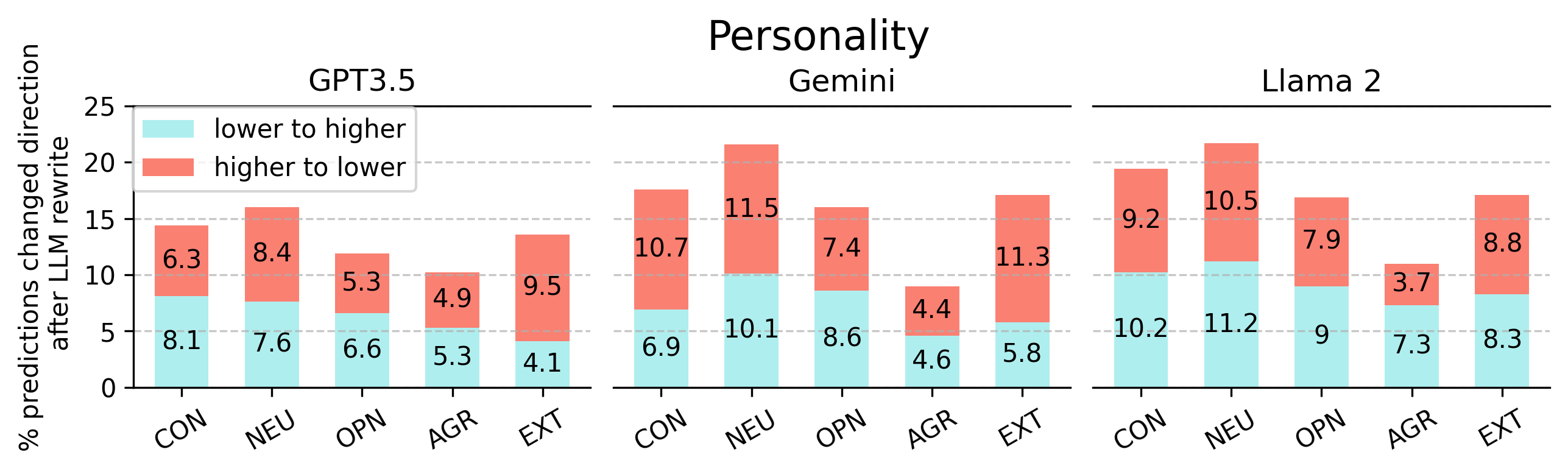}\hfill
\includegraphics[width=1.0\textwidth]{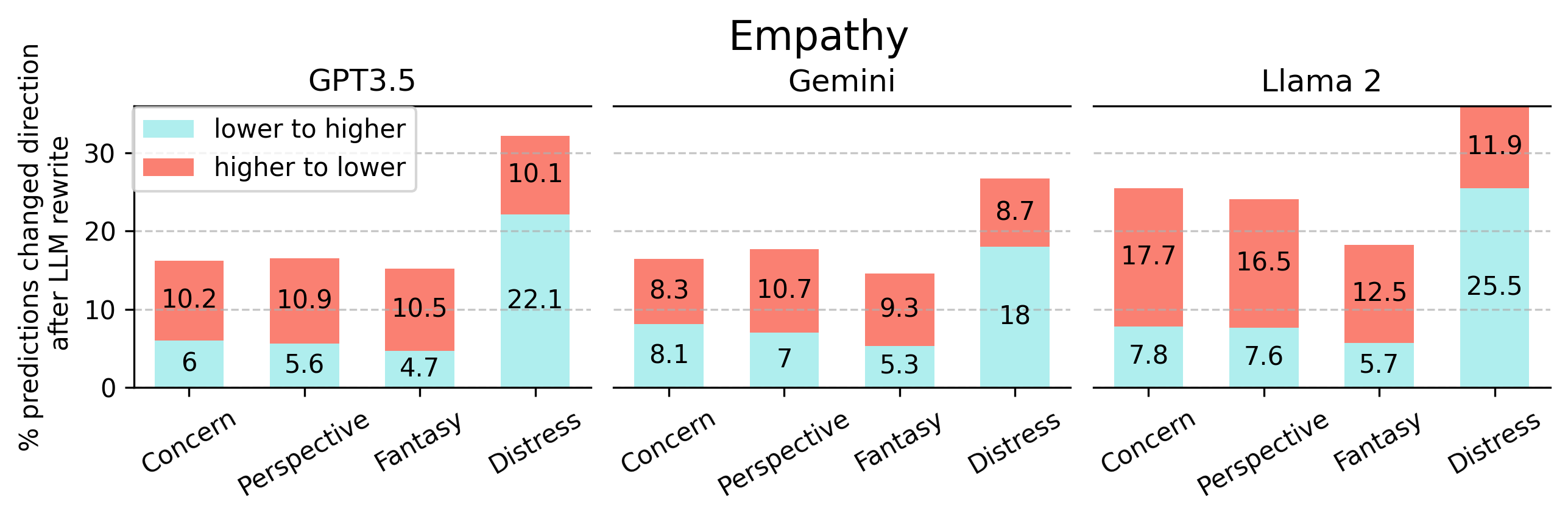}\hfill
\includegraphics[width=1.0\textwidth]{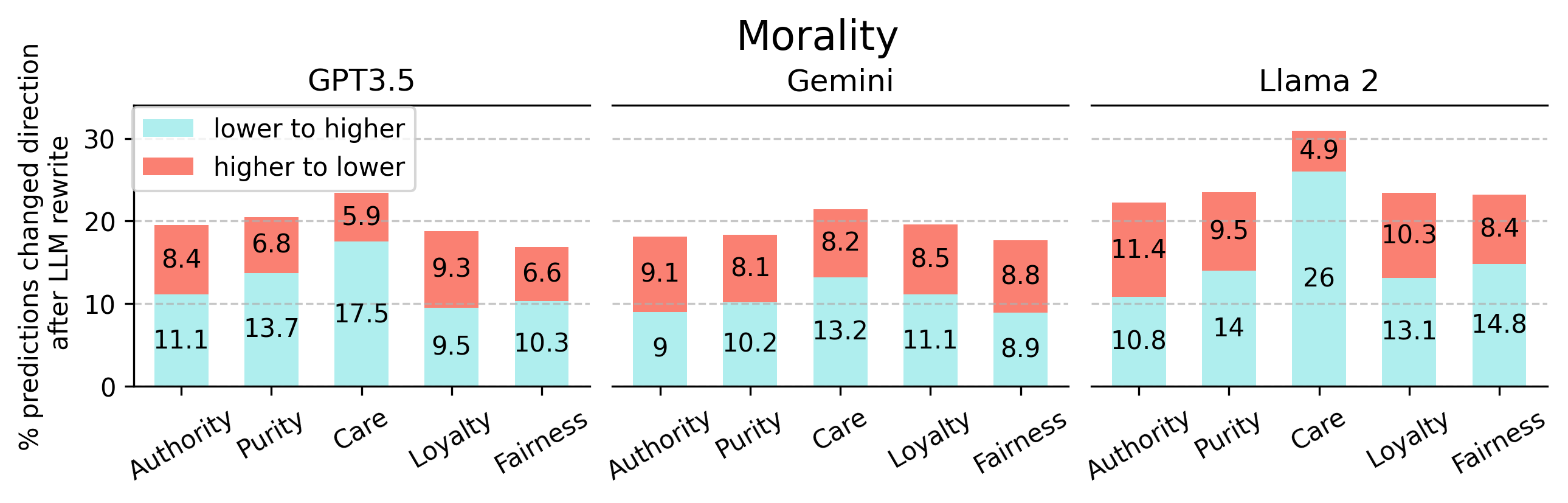}
\caption{Ratios of correct author attribute predictions on original texts that had different predicted labels on LLM-generated texts, grouped by the direction of change in predictions, broken down for different dimensions of investigated psychological constructs (i.e. personality, dispositional empathy, and morality).}
\label{fig:difference-in-predictions-with-all-dims}
\end{figure}

\section{What Author Attributes Do LLMs Promote?}
\label{appendix:What-Author-Attributes-do-LLMs-Promote-the-Most?}

In \autoref{do-llms-wash-away-stuff}, utilizing the direction of prediction changes, we iterated on the characteristics that LLMs promote in their own version of authors' texts. Expanding on that, in this section, we tried to provide a more fine-grained analysis for the investigated psychological constructs (i.e., personality, dispositional empathy, and morality) to contextualize our findings better with respect to the dimensions covered in each construct. \autoref{fig:difference-in-predictions-with-all-dims} demonstrates the ratio of predictions that changed from correct to incorrect, grouped by the direction of this change across different LLMs for personality, dispositional empathy, and morality dimensions. We only report on the trends with a Cohen's d effect size higher than negligible ($|d|<0.2$). 

In the case of specific characteristics related to personality, we observed that LLM-generated text is associated with people with higher levels of openness ($t(1074) = 6.50, p < .001, d = 0.21$ [small effect size]), agreeableness ($t(1005) = 9.05, p < .001, d = 0.27$ [small effect size]), and lower levels of extraversion ($t(1054) = 15.14, p < .001, d = 0.52$ [medium effect size]) compared to the actual authors. 

In the case of specific characteristics related to dispositional empathy, we observed that LLM-generated text is associated with people with higher levels of personal distress ($t(319) = 3.90, p < .001, d = 0.36$ [small effect size]), and lower levels of empathetic concern ($t(1945) = 10.6, p < .001, d = 0.35$ [small effect size]), perspective-taking ($t(2341) = 14.17, p < .001, d = 0.42$ [small effect size]), and fantasy ($t(2090) = 14.00, p < .001, d = 0.42$ [small effect size]) compared to the actual authors. 

Finally, in the case of specific characteristics related to morality, we observed that LLM-generated text is associated with people with higher levels of fairness ($t(158) = 4.34, p < .001, d = 0.53$ [medium effect size]), care ($t(734) = 25.44, p < .001, d = 1.54$ [large effect size]), purity ($t(711) = 9.90, p < .001, d = 0.59$ [medium effect size]), and loyalty ($t(752) = 5.51, p < .001, d = 0.29$ [small effect size]) compared to the actual authors.

\end{document}